\documentclass[lettersize,journal]{IEEEtran}
\usepackage{amsmath,amsfonts}
\usepackage{algorithmic}
\usepackage{algorithm}
\usepackage{array}
\usepackage[caption=false,labelfont=sf,textfont=sf]{subfig}
\usepackage{textcomp}
\usepackage{stfloats}
\usepackage{url}
\usepackage{verbatim}
\usepackage{graphicx}
\usepackage{cite}
\usepackage[english]{babel}
\usepackage{soul,color}
\soulregister\cite7
\soulregister\ref7
\soulregister\ac7
\soulregister\acp7
\soulregister\underline7

\usepackage{threeparttable}
\usepackage{url}
\usepackage{cite}
\DeclareMathAlphabet{\pazocal}{OMS}{zplm}{m}{n}
\usepackage{amssymb}
\usepackage{nicefrac}
\usepackage{optidef}
\usepackage{gensymb}
\usepackage{hyperref}
\usepackage{xurl}
\usepackage{orcidlink}
\usepackage{booktabs}
\usepackage{xcolor}

\usepackage{tikz}
\usetikzlibrary{shapes.geometric, arrows.meta, positioning, fit, backgrounds, calc}

\begin{document}

\title{Learning to Reflect: Hierarchical Multi-Agent Reinforcement Learning for CSI-Free mmWave Beam-Focusing}

\author{Hieu Le, \textit{Texas A\&M University},  Oguz Bedir, \textit{Texas A\&M University}, Jian Tao, \textit{Texas A\&M University}, Mostafa Ibrahim, \textit{Texas A\&M University}, and Sabit Ekin, \textit{Texas A\&M University}
        % <-this % stops a space
% \thanks{Manuscript received April 19, 2021; revised August 16, 2024.}
}

\markboth{Journal of \LaTeX\ Class Files,~Vol.~14, No.~8, August~00}%
{Shell \MakeLowercase{\textit{et al.}}: A Sample Article Using IEEEtran.cls for IEEE Journals}

% \IEEEpubid{0000--0000/00\$00.00~\copyright~2024 IEEE}

\maketitle

\begin{abstract}

Reconfigurable Intelligent Surfaces promise to transform wireless environments, yet practical deployment is hindered by the prohibitive overhead of Channel State Information (CSI) estimation and the dimensionality explosion inherent in centralized optimization. This paper proposes a Hierarchical Multi-Agent Reinforcement Learning (HMARL) framework for the control of mechanically reconfigurable reflective surfaces in millimeter-wave (mmWave) systems. We introduce a "CSI-free" paradigm that substitutes pilot-based channel estimation with readily available user localization data. To manage the massive combinatorial action space, the proposed architecture utilizes Multi-Agent Proximal Policy Optimization (MAPPO) under a Centralized Training with Decentralized Execution (CTDE) paradigm. The proposed architecture decomposes the control problem into two abstraction levels: a high-level controller for user-to-reflector allocation and decentralized low-level controllers for low-level focal point optimization. Comprehensive ray-tracing evaluations demonstrate that the framework achieves 2.81–7.94~dB RSSI improvements over centralized baselines, with the performance advantage widening as system complexity increases. Scalability analysis reveals that the system maintains sustained efficiency, exhibiting minimal per-user performance degradation and stable total power utilization even when user density doubles. Furthermore, robustness validation confirms the framework's viability across varying reflector aperture sizes (45–99 tiles) and demonstrates graceful performance degradation under localization errors up to 0.5~m. By eliminating CSI overhead while maintaining high-fidelity beam-focusing, this work establishes HMARL as a practical solution for intelligent mmWave environments.

\end{abstract}

\begin{IEEEkeywords}
Reconfigurable Intelligent Surfaces (RIS), Path Gain, Ray Tracing, Coverage Map, Deep Reinforcement Learning
\end{IEEEkeywords}

\section{Introduction}
\label{introduction}

The unprecedented surge in wireless traffic demand, fueled by augmented reality, autonomous systems, and massive deployments of IoT, has driven conventional wireless communication architectures to their theoretical limits. Traditional methodologies conceptualize the radio propagation medium as an immutable adversarial channel that degrades signal quality, necessitating progressively complex signal processing algorithms and elevated transmission power levels to compensate for channel degradation effects. This approach has resulted in performance saturation with a concurrent increase in power consumption and complexity of the system architecture \cite{direnzo:2020}.

Reconfigurable intelligent surfaces (RIS) transform previously passive structural elements into dynamic controllers of electromagnetic wave propagation, enabling adaptive radio environments. Nevertheless, despite the significant theoretical potential, practical RIS deployments face substantial implementation challenges. The primary obstacle involves the computational burden of channel state information (CSI) estimation, which requires accurate electromagnetic characterization across hundreds to thousands of reflecting components operating simultaneously. This requirement generates processing overhead that increases exponentially with architectural scale \cite{bjornson2022reconfigurable}.

Moreover, RIS methodologies depend extensively on constructive wave interference optimization, requiring advanced hardware components such as precision phase-shifting devices and ultra-low-latency reconfiguration systems. These demanding specifications, coupled with the mandatory perfect temporal synchronization in all reflecting elements, have inhibited large-scale commercial adoption \cite{pan2022overview, kim2022practical, a9864655}.

We present a fundamentally different methodology that bypasses the CSI estimation requirements through operation at a higher conceptual abstraction. In this context, we define "CSI-free" as the elimination of pilot-based electromagnetic channel estimation. We explicitly trade the prohibitive computational burden of high-dimensional channel estimation for a dependency on user localization, leveraging the fact that positioning data is more accessible and scalable than per-element channel estimation in large-scale RIS deployments. Rather than coordinating precise electromagnetic interference phenomena, our technique exploits spatial awareness and user positioning data to enhance reflection characteristics via macro-scale propagation management in non-line-of-sight (NLOS) scenarios.

\subsection{Hierarchical Multi-Agent Reinforcement Learning Paradigm}

The main innovation reformulates reflector optimization as a Hierarchical Multi-Agent Reinforcement Learning (HMARL) framework \cite{makar2001hierarchical} utilizing Centralized Training with Decentralized Execution (CTDE) \cite{kraemer2016multi}. To implement this, we employ Multi-Agent Proximal Policy Optimization (MAPPO) \cite{yu2022surprising}, which ensures stable cooperative learning by addressing the non-stationarity inherent in multi-agent environments. In contrast to traditional RIS implementations based on passive metamaterials, our framework controls mechanically reconfigurable metallic reflectors. While this introduces mechanical constraints, it offers distinct advantages: the elimination of complex RF circuitry, extended operational bandwidth, and streamlined control via conventional servos. Practical viability is further supported by the increasing availability of deep learning acceleration in modern communication platforms \cite{nasari2022benchmarking, le2024insight, qualcomm2024unlocking}.

To manage the complexity of multi-reflector coordination, the framework decomposes the control problem into two abstraction levels. At the high level, a coordinating controller performs intelligent user-to-reflector assignment based on spatial positioning. At the low level, specialized controllers autonomously optimize focal points for their assigned users. This hierarchical decomposition reduces observation spaces for efficient learning and employs temporal abstraction to balance long-term planning with rapid local adaptation.

By integrating spatial intelligence, this architecture effectively bypasses the CSI estimation bottleneck \cite{bjornson2022reconfigurable}. The system operates on user localization data rather than electromagnetic precision, achieving substantial performance improvements through large-scale propagation control. This approach ensures computational tractability and enables effective scaling across varying user densities and reflector configurations, maintaining optimal performance without explicit inter-agent data exchange.

\subsection{Contributions and Paper Organization}

This research delivers three primary contributions:

\begin{itemize}
    \item \textbf{NLOS CSI-free operation with substantial RSSI gains:} We formulate reflective surface optimization as a hierarchical multi-agent Markov Decision Process (HMA-MDP). This approach enables effective radio propagation management using only user localization data, eliminating the dependency on pilot-based channel estimation. The framework achieves Received Signal Strength Indicator (RSSI) improvements of 2.81--7.94~dB over centralized optimization baselines.
    
    \item \textbf{Scalable hierarchical allocation strategy:} We develop a two-level neural architecture comprising a high-level allocator and low-level controllers. This decomposition exhibits superior scalability: doubling the user density (from 2 to 4 users) results in a marginal 1.39~dB per-user performance degradation.
    
    \item \textbf{Comprehensive validation of hardware and algorithmic robustness:} We validate the framework's practical viability across diverse deployment conditions, including varying reflector aperture sizes (45--99 tiles), reward function formulations, and localization error levels. The system demonstrates high resilience, maintaining performance stability (reward standard deviation $\le$ 0.81~dBm) and graceful degradation under localization errors up to 0.5~m, confirming its suitability for real-world deployment without scenario-specific tuning.
\end{itemize}

This work on the hierarchical multi-agent approach is particularly well-suited for practical wireless systems that require reliable coverage in demanding environments. Key applications include indoor millimeter-wave (mmWave) communications and dense urban deployments. The framework excels at coordinating multiple reflecting elements by balancing global optimization with rapid local adaptation to changing propagation conditions. Our implementation is available on GitHub at \url{https://github.com/hieutrungle/rs}.

The remainder of this study is organized as follows: Section~\ref{sec:review} presents the literature survey. Section~\ref{sec:system_model} details the system architecture and mathematical formulation. Section~\ref{sec:marl_framework} describes the multi-agent reinforcement learning framework. Section~\ref{sec:results_and_discussion} examines simulation outcomes. Lastly, Section~\ref{sec:conclusion} summarizes primary findings and identifies future research directions.

\section{Literature Review}
\label{sec:review}

The field of RIS technology has progressed rapidly in recent years, driven by the increase in wireless traffic demands and the fundamental limitations of conventional communication architectures. Conventional RIS designs primarily leveraged electronically controlled phase adjustments to induce constructive interference at receivers and thereby enhance data rates \cite{direnzo:2020, zahra:2021}. These approaches, however, depend critically on accurate CSI for each individual reflecting unit, a requirement that scales unfavorably with system size. As deployments grow to hundreds or even thousands of elements, the CSI estimation burden becomes the dominant impediment to practical large-scale realization \cite{basharat:2022}.

In RIS-assisted wireless systems, obtaining accurate CSI typically requires a sequence of RIS reconfiguration states, leading to pilot overhead that scales with both the number of reflecting elements and the number of active users. Conventional cascaded channel estimation schemes therefore incur pilot lengths on the order of hundreds or even thousands of symbols, which causes significant spectral efficiency loss and, in fast-fading environments, estimation delays that exceed the channel coherence time \cite{a9400843}. To mitigate these burdens, several enhanced strategies have been investigated, including ON/OFF-based training protocols \cite{a10053657}, Discrete Fourier Transform (DFT)-structured estimation approaches \cite{a9328501}, and compressive sensing frameworks that exploit the inherent sparsity of mmWave channels \cite{a10016718}. Nevertheless, key challenges persist, particularly the high dimensionality of cascaded channel representations, limited scalability as system size grows, and constraints imposed by practical RIS hardware.

Statistical-CSI–based RIS architectures have demonstrated that effective beamforming can be realized without relying on instantaneous per-element channel estimates. Multi-port network models treat the RIS as a bidirectional scattering system, where statistical CSI enables improvements in average throughput through eigenmode selection and covariance-driven phase design \cite{a10666709}. In parallel, blind beamforming strategies eliminate explicit CSI estimation by iteratively updating the RIS phase shifts using only received signal strength feedback, achieving an $O(N^{2})$ SNR scaling for an $N$-element surface across diverse propagation conditions \cite{lai2023blind}. Codebook-driven designs further reduce signaling overhead by employing pre-optimized discrete phase-shift sets, with well-constructed codebooks approaching the performance of full-CSI schemes while requiring minimal feedback \cite{a9952197}. Additionally, location-aware RIS techniques leverage user position information in the absence of full CSI, integrating localization and beamforming through optimization criteria grounded in the Cramer–Rao bound \cite{nazar2024revolutionizing}.

The adoption of deep reinforcement learning (DRL) for RIS configuration has aimed to address the limitations of traditional optimization schemes. Early contributions demonstrated sum-rate maximization frameworks that relied on full CSI availability across all relevant links \cite{huang2020reconfigurable}, while subsequent studies proposed DRL-based phase shift design leveraging discrete channel vector samples \cite{taha2020deep, taha2021enabling}. More recent work integrates sensing hardware within the RIS to facilitate distributed channel estimation prior to DRL-driven optimization \cite{choi2024deep}. Nonetheless, these approaches still require explicit channel estimation at the RIS, imposing substantial increases in hardware complexity and power consumption. Recent initiatives to relax CSI dependence have yielded mixed results \cite{sheen2021deep}, as many such solutions rely on large offline training datasets that limit adaptability in rapidly varying environments.

The transition toward multi-agent reinforcement learning (MARL) for RIS optimization marks a significant advancement in addressing the coordination challenges associated with large-scale intelligent surface control. Recent MARL-based implementations have demonstrated substantial performance gains; for example, Multi-Agent Twin Delayed Deep Deterministic Policy Gradient (TD3) schemes applied to joint beamforming and RIS codebook design achieve performance on par with 256-beam DFT codebooks while reducing training computation by 97\% \cite{a10060056}. Similarly, the Multi-Agent Global and locAl deep Reinforcement learning (MAGAR) framework \cite{a10758034} for STAR-RIS systems yields an 18\% improvement in energy efficiency over conventional MARL , highlighting the potential of distributed learning mechanisms for RIS management. Moreover, authors of \cite{aa11322690} demonstrated that MARL framework can achieve around 5 dB higher in average RSSI for multiple users over single-agent DRL framework.

MARL methods can be characterized according to the nature of agent interactions: fully cooperative, fully competitive, or mixed. In fully cooperative settings, all agents jointly pursue a unified long-term objective and potentially share a common reward signal \cite{busoniu2008comprehensive, zhang2018fully}. This category includes models such as multi-agent Markov decision processes and team-average reward formulations \cite{kar2013cal, zhang2018fully}. Competitive MARL corresponds to zero-sum Markov games \cite{littman1994markov, shapley1953stochastic, park2023multi}, whereas mixed scenarios integrate elements of both cooperation and competition \cite{hu2003nash, lowe2017multi}.

HMARL has emerged as an effective framework for partitioning intricate coordination tasks into multiple layers of abstraction. Foundational studies extended hierarchical RL constructs such as MAXQ \cite{maxq1998} to multi-agent domains \cite{ghavamzadeh2006hierarchical, makar2001hierarchical}, enabling agents to acquire high-level coordination strategies while executing primitive behaviors at lower tiers. More recent advancements include ALlocator-Actor Multi-Agent Architecture (ALMA) \cite{iqbal2022alma}, which learns subtask allocation policies in tandem with low-level execution policies, and Hierarchical Multi-Agent Skill Discovery (HMASD) \cite{yang2023hierarchical}, which utilizes transformer-based architectures for sequential skill assignment. The efficacy of hierarchical techniques stems from their ability to exploit temporal abstraction for coordinated decision-making while retaining scalability through modular policy designs that extend across heterogeneous agent populations and dynamic environments \cite{xu2023haven}.

Although extensive research has centered on electronically controlled RIS platforms, mechanically reconfigurable metallic reflectors represent a fundamentally different class of intelligent surfaces that has received comparatively limited study. Unlike electronic RIS architectures that require complex RF circuitry, metallic reflectors inherently offer wideband operation, e.g. almost frequency agnostic, simplified actuation via conventional servo systems, and elimination of stringent electromagnetic tuning requirements. Recent studies on passive metallic reflectors at 28 GHz have shown that appropriately positioned flat reflectors can yield substantial coverage and gain improvements in non-line-of-sight scenarios without introducing electronic complexity \cite{a8972365, a9500547, le2024guiding}. Complementary theoretical investigations have proposed enhanced reflection models that highlight the breakdown of classical Snell's law when reflector dimensions approach the signal wavelength \cite{a10279522}. While active or amplifier-equipped RIS designs can substantially increase link capacity \cite{a9998527}, they also incur additional hardware complexity, power consumption, and noise amplification—limitations inherently avoided by passive mechanically reconfigurable metallic reflectors.

The integration of MARL with mechanically reconfigurable metallic reflectors defines a potential research direction that addresses critical gaps in existing intelligent surface technologies. This paradigm enables the avoidance of CSI estimation requirements that burden traditional RIS systems, while still delivering meaningful performance gains through large-scale geometric propagation control rather than fine-grained electromagnetic manipulation. We introduce a CSI-free framework that exploits spatial awareness and user location information to determine reflector configurations at a higher abstraction level. The proposed hierarchical control architecture comprises two synergistic components: a high-level allocation module responsible for selecting optimal reflector segments for each user, and low-level policies that refine reflector orientations to maximize RSSI for their assigned users. This hierarchical design effectively decomposes the underlying multi-objective optimization problem, enhancing scalability and adaptability across heterogeneous deployment environments.

\section{SYSTEM MODEL AND PROBLEM FORMULATION}
\label{sec:system_model}

\subsection{System Architecture and Problem Formulation}
\label{subsec:system_architecture}

The reflective surface optimization problem exhibits high computational complexity due to the joint user-to-reflector assignment and continuous reflector configuration challenges. To address this complexity while maintaining practical deployability, we employ a hierarchical control architecture that decomposes the system-wide optimization into high-level allocation and low-level execution subproblems. This section establishes the physical system model, introduces the hierarchical decomposition rationale, and formalizes the constraints governing reflector operation.

\subsubsection{Physical System Configuration}

\begin{figure}[!htp]
    \centerline{\includegraphics[width=1.0\linewidth]{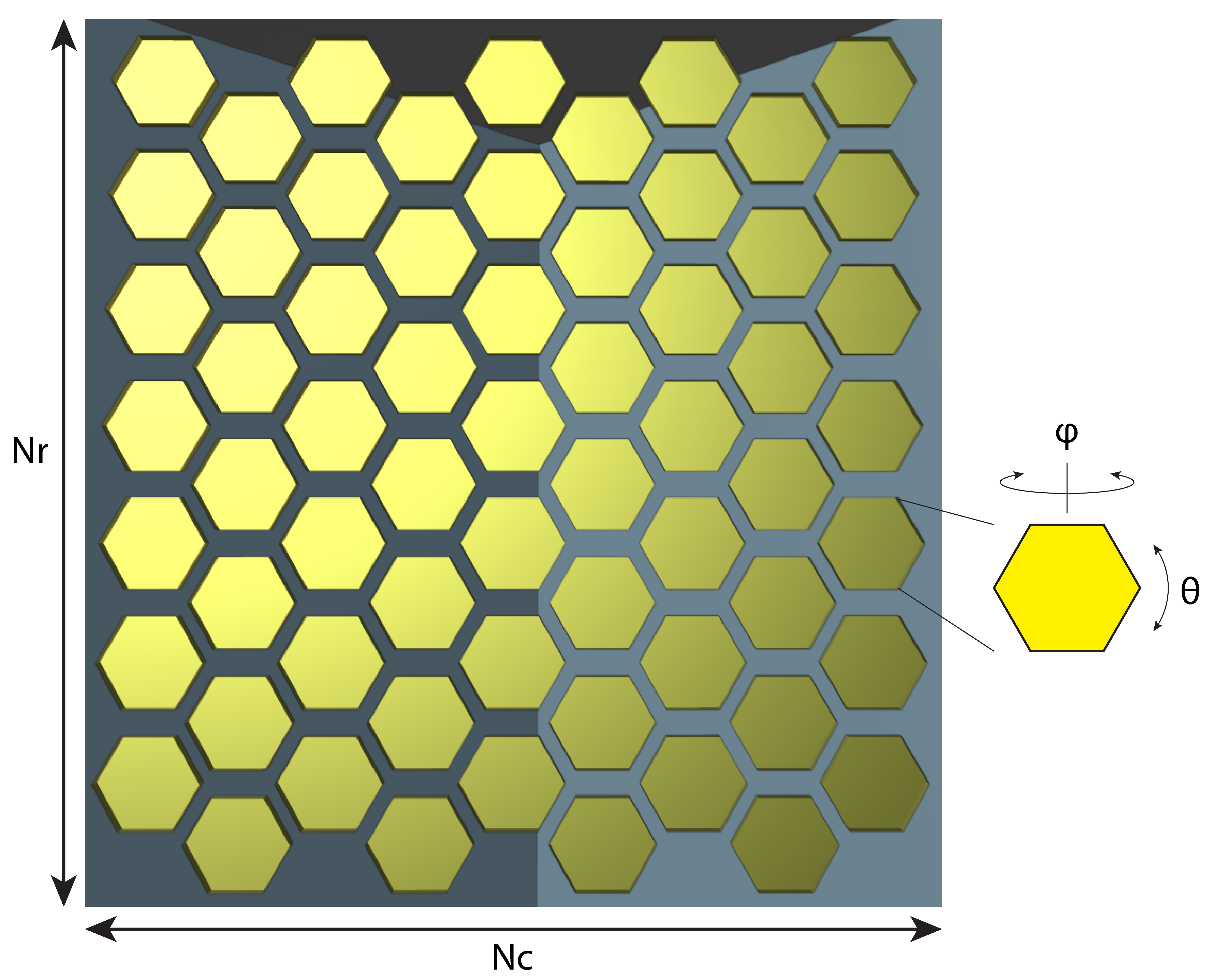}}
    \caption{Reflector with hexagonal tiles. Each tile can be rotated in both elevation ($\theta$) and azimuth ($\phi$) angles. \label{Figure:reflector}}
    % \label{Figure:reflector}
\end{figure}

The reflective surface system comprises three primary components: an access point (AP), $K$ user equipment (UE) devices, and $L$ independently controllable reflector segments. The reflector array consists of $N_r \times N_c$ hexagonal metallic tiles arranged in a planar configuration as shown in Fig.~\ref{Figure:reflector}, where each tile provides mechanical adjustment capabilities in both elevation angle $\theta_{i,j}$ and azimuth angle $\phi_{i,j}$. This mechanical reconfigurability enables electromagnetic wavefront manipulation without requiring sophisticated RF circuitry or electronic phase shifters, making the system particularly suitable for mmWave and sub-terahertz frequencies where conventional phased arrays face significant implementation challenges.

The reflector arrays employ a hierarchical segmentation scheme where tiles are partitioned into $L$ spatial segments:
\begin{equation}
\mathcal{S} = \mathcal{S}_1 \cup \mathcal{S}_2 \cup \ldots \cup \mathcal{S}_L, \quad \mathcal{S}_i \cap \mathcal{S}_j = \emptyset, \, \forall i \neq j,
\end{equation}
where $\mathcal{S}_l = \{(i,j) : \text{tile } (i,j) \text{ belongs to segment } l\}$ denotes the set of tiles within segment $l$. Each segment $l$ is characterized by its centroid position $r_l \in \mathbb{R}^3$ and controls a focal point $f_l(t) \in \mathbb{R}^3$ that governs the collective reflection behavior of all tiles within that segment. The focal point concept provides an abstraction that reduces the control complexity from $2N_rN_c$ individual tile parameters to $3L$ segment-level focal point coordinates, where typically $L \ll N_rN_c$.

UE positions are denoted as $u_k(t) = [u_{k,x}(t), u_{k,y}(t), u_{k,z}(t)]^\top \in \mathbb{R}^3$ for $k \in \{1, 2, \ldots, K\}$, and the access point is located at fixed position $s \in \mathbb{R}^3$. The system operates in time-division duplex (TDD) mode with discrete time steps indexed by $t \in \{0, 1, 2, \ldots\}$.

\subsubsection{Hierarchical Control Architecture}

The computational complexity of jointly optimizing user-to-reflector assignments and individual reflector tile configurations, even with angle quantization assumption, scales as $\mathcal{O}(K^L \times (\mathcal{Q}_r \times \mathcal{Q}_c)^{N_r \times N_c})$, rendering exhaustive search intractable for practical system sizes. $\mathcal{Q}_r$ and $\mathcal{Q}_c$ represent the quantization levels of azimuth and elevation angular steps, respectively. To achieve scalable optimization, we decompose the control problem into a two-tier hierarchical architecture that exploits the composite task structure inherent in wireless communication scenarios.

\textbf{High-level Allocation Layer:} The high-level controller operates as a centralized decision-maker responsible for high-level user-to-reflector segment assignment. 
This controller determines the allocation $b = \{b_1, b_2, \ldots, b_L\}$, where $b_l \in \{1, 2, \ldots, K\}$ indicates that user $k$ is assigned to reflector segment $b_l$. 
The allocation action space $\mathcal{B}$ contains $K^L$ possible assignments, representing a discrete combinatorial optimization problem. Critically, the high-level controller operates on a temporally extended time scale, making allocation decisions every $T$ environment time steps. This temporal abstraction allows sufficient time for lower-level controllers to optimize their focal point configurations before high-level reassignment occurs.

\textbf{Low-level Execution Layer:} Given an allocation decision from the high-level controller, each reflector segment $l$ autonomously optimizes its focal point position $f_l(t)$ to maximize the received signal strength for its currently assigned user. These low-level controllers operate at every time step with decentralized execution, adjusting focal points based on local observations of their assigned user's position and current focal point state. The action space for each low-level controller is continuous, consisting of focal point displacements $a_{l,t} = [\Delta f_{l,x}, \Delta f_{l,y}, \Delta f_{l,z}]^\top \in \mathbb{R}^3$ subject to maximum displacement constraints.

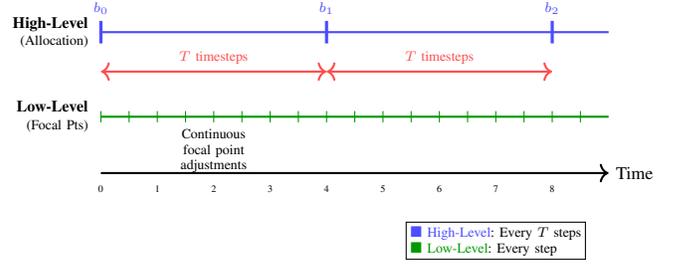
\begin{figure}[t]
\centering
\begin{tikzpicture}[
    scale=0.75,
    transform shape,
    every node/.style={font=\footnotesize}
]

% Time axis
\draw[thick, ->] (0,0) -- (9,0) node[right] {\small Time};

% Timestep markers
\foreach \x in {0,1,2,3,4,5,6,7,8} {
    \node[below] at (\x,-0.1) {\tiny \x};
}

% High-level timeline
\draw[thick, blue!70] (0,2.5) -- (9,2.5);
\node[left, align=right] at (-0.1,2.5) {\textbf{High-Level}\\ \scriptsize (Allocation)};

% High-level decisions (at t=0, t=4, t=8)
\foreach \x/\label in {0/$b_0$, 4/$b_1$, 8/$b_2$} {
    \draw[blue!70, very thick] (\x,2.3) -- (\x,2.7);
    \node[above, blue!70, font=\scriptsize] at (\x,2.7) {\label};
}

% Low-level timeline
\draw[thick, green!60!black] (0,1) -- (9,1);
\node[left, align=right] at (-0.1,1) {\textbf{Low-Level}\\ \scriptsize (Focal Pts)};

% Low-level actions (every timestep)
\foreach \x in {0,0.5,1,1.5,2,2.5,3,3.5,4,4.5,5,5.5,6,6.5,7,7.5,8,8.5} {
    \draw[green!60!black, thin] (\x,0.9) -- (\x,1.1);
}

% Temporal abstraction indicators
\draw[<->, thick, red!70] (0,1.8) -- (4,1.8) node[midway, above, font=\scriptsize] {$T$ timesteps};
\draw[<->, thick, red!70] (4,1.8) -- (8,1.8) node[midway, above, font=\scriptsize] {$T$ timesteps};

% Annotations
\node[align=center, font=\scriptsize] at (2,0.4) {Continuous\\focal point\\adjustments};

% Compact legend
\node[draw, rectangle, fill=white, align=left, font=\scriptsize, inner sep=2pt] at (7,-1.2) {
    \textcolor{blue!70}{$\blacksquare$ High-Level}: Every $T$ steps\\
    \textcolor{green!60!black}{$\blacksquare$ Low-Level}: Every step
};

\end{tikzpicture}
\caption{Temporal coordination between hierarchical controllers. The high-level allocation controller makes high-level user-to-reflector assignment decisions every $T$ timesteps, while low-level controllers continuously optimize focal points at every timestep. This temporal abstraction enables high-level planning at the high level while maintaining rapid low-level adaptation at the low level.}
\label{Figure:temporal_coordination}
\end{figure}

\textbf{Temporal Coordination Mechanism:} The hierarchical architecture employs temporal abstraction to coordinate decision-making across levels. Fig.~\ref{Figure:temporal_coordination} illustrates this coordination mechanism: the high-level controller makes allocation decisions at time steps $t \in \{0, T, 2T, \ldots\}$, while low-level controllers continuously optimize focal points at every time step. This structure enables high-level planning at longer time scales while maintaining rapid low-level adaptation to environmental dynamics.

The hierarchical decomposition provides three fundamental advantages for practical deployment. First, it achieves observation space reduction by enabling low-level controllers to operate on masked local observations relevant only to their assigned users, rather than requiring full system state knowledge. Second, it enables modular policy reuse where learned low-level focal point control policies can transfer across different user configurations and allocation decisions. Third, it provides computational tractability by separating the discrete combinatorial allocation problem from the continuous focal point optimization problem, enabling specialized solution techniques for each subproblem.

\subsubsection{Physical Constraints and Feasibility Conditions}

The hierarchical control architecture must respect physical limitations imposed by the mechanical reconfiguration mechanism and geometric constraints. These constraints are naturally integrated through the focal point abstraction.

\textbf{Tile Orientation Constraints:} For each tile $(i,j)$ belonging to segment $l$, the tile orientation is determined by the segment's focal point position $f_l(t)$ through geometric reflection principles. The tile normal vector is computed as:
\begin{equation}
\vec{n}_{i,j}(f_l) = \frac{1}{2}\left(\frac{f_l - r_{i,j}}{\|f_l - r_{i,j}\|_2} + \frac{s - r_{i,j}}{\|s - r_{i,j}\|_2}\right),
\end{equation}
where $r_{i,j} \in \mathbb{R}^3$ denotes the position of tile $(i,j)$. The elevation and azimuth angles are then derived as:
\begin{equation}
\theta_{i,j}(f_l) = \arccos(\vec{n}_{i,j}(f_l) \cdot \hat{z}),
\end{equation}
\begin{equation}
\phi_{i,j}(f_l) = \text{atan2}(\vec{n}_{i,j}(f_l) \cdot \hat{y}, \vec{n}_{i,j}(f_l) \cdot \hat{x}),
\end{equation}
where $\hat{z}$, $\hat{y}$, and $\hat{x}$ are unit vectors along the coordinate axes.

The mechanical actuation system imposes physical limits on achievable tile orientations:
\begin{equation}
\begin{split}
    \theta_{\min} \leq \theta_{i,j}(f_l) &\leq \theta_{\max}, \\
    \phi_{\min} \leq \phi_{i,j}(f_l) &\leq \phi_{\max}, \\
    & \quad \forall (i,j) \in \mathcal{S}_l.
\end{split}
\end{equation}
These constraints define the feasible focal point region $\mathcal{F}_l$ for segment $l$:
\begin{equation}
\begin{split}
\mathcal{F}_l = \Big\{ f \in \mathbb{R}^3 : \, & \theta_{\min} \leq \theta_{i,j}(f) \leq \theta_{\max}, \\
& \phi_{\min} \leq \phi_{i,j}(f) \leq \phi_{\max}, \, \forall (i,j) \in \mathcal{S}_l \Big\}.
\end{split}
\end{equation}

In practice, we approximate $\mathcal{F}_l$ using axis-aligned bounding boxes:
\begin{equation}
\mathcal{F}_l \approx \{f \in \mathbb{R}^3 : f_{l,\min} \leq f \leq f_{l,\max}\},
\end{equation}
where $f_{l,\min}$ and $f_{l,\max}$ are determined through geometric analysis of the segment's tile positions and mechanical constraints.

\textbf{Action Space Constraints:} The low-level controllers' focal point displacement actions are constrained to ensure smooth transitions and respect actuation limits:
\begin{equation}
\|a_{l,t}\|_\infty \leq \delta_{\max}, \quad f_l(t+1) = f_l(t) + a_{l,t} \in \mathcal{F}_l,
\end{equation}
where $\delta_{\max}$ defines the maximum focal point displacement per time step, typically set based on the mechanical actuation speed.

\subsubsection{Signal Propagation Model}

The controllable RSSI at user $k$ depends on both direct propagation from the access point and reflected paths through the assigned reflector segment. For user $k$ assigned to segment $l$ under allocation $b$, the received power is modeled as:
\begin{equation}
\begin{split}
P_{r,k}(u_k, f) &={} P_t \sum_{(i,j) \in \mathcal{S}_{b}} |h_{r,k}^{(i,j)}(u_k, f) + h_{\text{other},k}(u_k)|^2,
\label{eq:received_power}
\end{split}
\end{equation}
where $P_t$ denotes the transmit power, $h_{r,k}^{(i,j)}(u_k, f)$ represents total reflected channel coefficient from tile $(i,j)$ to user $k$ as a function of the user location and focal point position, and $h_{\text{other},k}(u_k)$ accounts for the other propagation path. These coefficients are obtained from a deterministic ray-tracing model of a fixed propagation environment, so for a given geometry they are fully determined by user and focal-point locations rather than by random fading. Since we focus on NLOS scenarios, no direct path exists.

The system-wide performance objective aggregates received power across all users:
\begin{equation}
R_{\text{sys}}(s(t), b(t)) = \sum_{k=1}^{K} P_{r,k}(u_k(t), f(t)).
\label{eq:system_reward}
\end{equation}

We emphasize that the system-level reward is a state-dependent function rather than purely time-dependent. While we index states by time $t$ to denote temporal evolution, the reward function $R_{\text{sys}}(s(t), b(t))$ itself depends on the current configuration of user positions, focal points, and allocations. This formulation aligns with standard MDP conventions where rewards are functions of states, and time dependence enters only through state evolution according to the system dynamics. This formulation also provides a differentiable performance metric that couples the high-level allocation decisions (through the assignment mapping $b_l$) with low-level focal point configurations (through $f_l(t)$), enabling coordinated hierarchical optimization.

\subsection{Computational Complexity Reduction}
The hierarchical architecture with focal point abstraction achieves dimensionality reduction compared to direct tile-level optimization. The control parameter dimension for the proposed focal-point-based approach scales as:
\begin{equation}
    \mathcal{D}_{\text{focal}} = |\mathcal{B}| + 3L = K^L + 3L,
\end{equation}
where the first term represents the discrete allocation space size and the second term represents the continuous focal point coordinates. In contrast, direct tile-level optimization requires a parameter space of:
\begin{equation}
    \mathcal{D}_{\text{tile}} = K^L + 2N_rN_c,
\end{equation}
since each tile requires two independent orientation angles ($\theta, \phi$). To rigorously quantify the reduction efficiency, we define the dimensionality reduction factor $\eta(K, L, N)$, assuming square reflector arrays where $N_r = N_c = N$:
\begin{equation}
    \eta(K, L, N) = \frac{\mathcal{D}_{\text{tile}}}{\mathcal{D}_{\text{focal}}} = \frac{K^L + 2N^2}{K^L + 3L}.
\end{equation}
We analyze the gradient of $\eta$ with respect to system parameters to identify the operating regimes where the hierarchical design is most effective.

\subsubsection{Sensitivity to Surface Resolution ($N$)}
Differentiating $\eta$ with respect to the number of tiles per dimension $N$:
\begin{equation}
    \frac{\partial \eta}{\partial N} = \frac{\partial}{\partial N} \left( \frac{K^L + 2N^2}{K^L + 3L} \right) = \frac{4N}{K^L + 3L}.
\end{equation}
Since $N > 0$, the gradient is always positive ($\frac{\partial \eta}{\partial N} > 0$). 
Because $\eta(N)$ contains a dominant $N^2$ term in the numerator, the reduction factor grows quadratically with $N$, and because $\frac{\partial \eta}{\partial N} > 0$ for $N>0$, it is strictly increasing in $N$. As reflector arrays become denser to support higher frequency beamforming, the complexity savings of the focal point abstraction become increasingly significant.

\subsubsection{Sensitivity to Reflector Segments ($L$)}
To determine how the reduction factor scales with the number of reflector segments $L$, we apply the quotient rule $\left(\frac{u}{v}\right)' = \frac{u'v - uv'}{v^2}$. Let $u = K^L + 2N^2$ and $v = K^L + 3L$. The derivatives with respect to $L$ are:
\begin{equation}
    \frac{\partial u}{\partial L} = K^L \ln K, \quad \frac{\partial v}{\partial L} = K^L \ln K + 3.
\end{equation}
Substituting these into the quotient rule yields:
% \begin{equation}
% \begin{aligned}
%     \frac{\partial \eta}{\partial L} &= \frac{(K^L \ln K)(K^L + 3L) - (K^L + 2N^2)(K^L \ln K + 3)}{(K^L + 3L)^2}
%     % &= \frac{(K^{2L} \ln K + 3L K^L \ln K) - (K^{2L} \ln K + 3K^L + 2N^2 K^L \ln K + 6N^2)}{(K^L + 3L)^2}.
% \end{aligned}
% \end{equation}
% Canceling the common term $K^{2L} \ln K$ and grouping terms by $K^L$, we obtain:
\begin{equation}
    \frac{\partial \eta}{\partial L} = \frac{K^L [ \ln K (3L - 2N^2) - 3 ] - 6N^2}{(K^L + 3L)^2}.
\end{equation}
In practical mmWave configurations, the hardware complexity term $2N^2$ typically dominates the segment count $3L$ (i.e., $2N^2 \gg 3L$). Consequently, the term $(3L - 2N^2)$ is negative, rendering the entire gradient negative ($\frac{\partial \eta}{\partial L} < 0$). This confirms that increasing $L$ reduces the compression ratio, as it introduces additional dimensions to the focal point abstraction (both continuous coordinates and allocation combinations).

\subsubsection{Sensitivity to User Density ($K$)}
Similarly, the sensitivity with respect to user density is derived as:
\begin{equation}
    \frac{\partial \eta}{\partial K} = \frac{L K^{L-1}(3L - 2N^2)}{(K^L + 3L)^2}.
\end{equation}
Under the same condition $2N^2 \gg 3L$, this derivative is negative, implying that the relative efficiency gain decreases as the allocation space $K^L$ expands.

\subsubsection{Complexity Regimes and Practical Implications}
The analytical behavior of $\eta$ highlights two distinct operating regimes based on the magnitude of $K$ and $L$:

\textbf{1. The Asymptotic Limit (Saturated Regime):}
In scenarios where the number of users $K$ or segments $L$ becomes very large (e.g., massive crowd connectivity), the combinatorial allocation term $K^L$ grows exponentially and dominates both the numerator and denominator ($K^L \gg 2N^2$). In this limit:
\begin{equation}
    \lim_{K, L \to \infty} \eta(K, L, N) \approx \frac{K^L}{K^L} = 1.
\end{equation}
Here, the complexity is bottlenecked by the discrete assignment problem, which affects both the proposed and baseline methods equally, negating the advantage of the focal point abstraction.

\textbf{2. The Practical Indoor Regime (Efficient Regime):}
Indoor mmWave deployments are naturally constrained by physical geometry and coverage requirements, placing them in a regime where the proposed method excels.
\begin{itemize}
    \item \textbf{Small $K$:} Room-scale coverage typically involves a limited number of active users requiring high-gain tracking (e.g., $K \in [1, 2]$).
    \item \textbf{Small $L$:} The number of reflector arrays is limited by installation costs of individual control chains (e.g., $L \in [1, 4]$).
    \item \textbf{Large $N$:} To overcome severe mmWave path loss, reflector arrays must be electrically large with high tile counts (e.g., $N \ge 8$, resulting in $\ge 64$ tiles per array).
\end{itemize}
In this regime, the condition $2N^2 \gg K^L$ holds strictly. The focal point abstraction compresses the massive hardware state space ($2N^2$) into a compact representation while the allocation overhead ($K^L$) remains manageable. For the specific configuration used in this study ($L=4, N=8, K=2$):
\begin{equation}
    \eta \approx \frac{16 + 128}{16 + 12} = \frac{144}{28} \approx 5.14.
\end{equation}
This represents a greater than five-fold reduction in the action space dimensionality, accelerating the convergence of the MARL agents.

% \textbf{Computational Complexity Reduction:} The hierarchical architecture with focal point abstraction achieves better complexity reduction compared to direct tile-level optimization. The control parameter dimension scales as:
% \begin{equation}
% \mathcal{D}_{\text{focal}} = |\mathcal{B}| + 3L = K^L + 3L,
% \end{equation}
% where the first term represents the discrete allocation space and the second term represents the continuous focal point coordinates. In contrast, direct tile-level optimization requires:
% \begin{equation}
% \mathcal{D}_{\text{tile}} = K^L + 2N_rN_c,
% \end{equation}
% since each tile has two orientation angles. For typical system configurations where $L = 4$, $N_r = N_c = 8, K = 2$, this yields a reduction factor of:
% \begin{equation}
% \frac{K^L + 2N_rN_c}{K^L + 3L} = \frac{K^L + 128}{K^L + 12} \approx 5.
% \end{equation}

% The hierarchical architecture and physical system model established in this section provide the foundation for the multi-agent reinforcement learning formulation presented in Section~III-B, where we formalize the control problem as a hierarchical Markov decision process and develop the learning framework for CSI-free optimization.

\subsection{Hierarchical Multi-Agent MDP Formulation}
\label{subsec:hma_formulation}

Building upon the physical system architecture and hierarchical control structure established in Section~\ref{subsec:system_architecture}, we now formalize the reflective surface optimization as a Hierarchical Multi-Agent Markov Decision Process (HMA-MDP). This formulation provides a mathematical framework for learning-based optimization without requiring CSI, leveraging the high-level-low-level decomposition to achieve computational tractability.

\subsubsection{High-Level Allocation MDP}

The high-level controller addresses the high-level problem of user-to-reflector segment assignment, operating as a centralized decision-maker with access to global system state. This level makes discrete combinatorial decisions on a temporally extended time scale.

\textbf{State Space:} The high-level state $s_H(t) \in \mathcal{S}_H$ encompasses global system information necessary for informed allocation decisions:
\begin{equation}
s_H(t) = \{u_1(t), \ldots, u_K(t), r_1, \ldots, r_L, f_1(t), \ldots, f_L(t)\}
\label{eq:high_level_state}
\end{equation}
where $u_k(t) = [u_{k,x}(t), u_{k,y}(t), u_{k,z}(t)]^\top \in \mathbb{R}^3$ represents the three-dimensional position of user $k$ at time $t$, and $r_l \in \mathbb{R}^3$ denotes the fixed centroid position of reflector segment $l$. The state space dimensionality scales as $|\mathcal{S}_H| = \mathbb{R}^{3K+6L}$, providing complete observability of user and reflector positions while abstracting away the detailed tile-level configuration.

\textbf{Action Space:} The allocation action space $\mathcal{A}_H = \mathcal{B}$ consists of all feasible assignments of users to reflector segments. An allocation $b \in \mathcal{B}$ is represented as:
\begin{equation}
\mathbf{b}(t) = [b_1(t), \ldots, b_L(t)]^\top , \quad b_l \in \{1, 2, \ldots, K\},
\label{eq:allocation_action}
\end{equation}
where $b_l = k$ indicates that user $k$ is assigned to reflector segment $l$. The cardinality of the action space is $|\mathcal{B}| = K^L$, representing a substantial combinatorial optimization challenge. For practical systems with $K=2$ users and $L=4$ reflector segments, this yields 128 possible allocations, necessitating efficient exploration strategies.

\textbf{Temporal Abstraction:} The high-level controller operates with temporal abstraction at time scale $T$, making allocation decisions at time steps $t \in \{0, T, 2T, \ldots\}$. The allocation $b(t_H)$ determined at high-level time $t_H$ remains fixed for the subsequent $T$ environment steps, allowing low-level controllers sufficient optimization horizon before high-level reassignment. This temporal commitment mechanism is critical for hierarchical coordination, as premature reassignment would destabilize low-level learning and prevent effective focal point convergence. Therefore, high-level controller has a much slower update rate than the low-level controller to stabilize the training.

\textbf{Value Function and Optimization Objective:} The high-level controller learns a state-action value function $Q_H: \mathcal{S}_H \times \mathcal{B} \rightarrow \mathbb{R}$ that estimates the expected cumulative reward for executing allocation $b$ in state $s_H$:
\begin{equation}
Q_H(s_H, b) = \mathbb{E}\left[\sum_{t=0}^{T-1} \gamma^t R_{\text{sys}}(s(t), b) \mid s(0) = s_H\right]
\label{eq:high_level_Q}
\end{equation}
where $\gamma \in (0,1)$ is the discount factor and $R_{\text{system}}$ is the system-wide reward defined in Equation~\eqref{eq:system_reward}. The high-level optimization objective seeks the allocation policy $\pi_H: \mathcal{S}_H \rightarrow \mathcal{B}$ that maximizes expected long-term performance:
\begin{equation}
\pi_H^* = \mathop{{\arg\max}}_{\pi_H} \mathop{\mathbb{E}}_{s_H \sim \rho, b \sim \pi_H(\cdot|s_H)}\left[\sum_{t=0}^{\infty} \gamma^t R_{\text{sys}}(s_H(t), b(t))\right],
\label{eq:high_level_objective}
\end{equation}
where $\rho$ denotes the state distribution induced by user mobility dynamics.

Due to the massive combinatorial action space, direct enumeration of all allocations is computationally prohibitive. We address this challenge through an amortized Q-learning approach with a learned proposal distribution $\pi_H(b|s_H; \zeta)$ parameterized by neural network weights $\zeta$, enabling efficient sampling of high-value allocations during both training and deployment.

\subsubsection{Low-Level Focal Point MDP}

Given an allocation $b$ from the high-level controller, each reflector segment $l \in \{1, 2, \ldots, L\}$ operates as an independent agent responsible for optimizing its focal point position to maximize signal strength for its assigned user. These low-level controllers operate at every environment time step with decentralized execution.

\textbf{State Space with Observation Masking:} Each low-level agent $l$ observes only information relevant to its assigned subtask, implementing the observation masking principle that is fundamental to scalable multi-agent learning. The local state $s_{L,l}(t) \in \mathcal{S}_{L,l}$ is defined as:
\begin{equation}
s_{L,l}(t) = \{u_{\pi(l)}(t), r_l, f_l(t)\},
\label{eq:low_level_state}
\end{equation}
where $\pi(l)$ denotes the user assigned to reflector segment $l$ under the current allocation $b$ (i.e., $\pi(l) = k$ if $b_l = k$), and $f_l(t) = [f_{l,x}(t), f_{l,y}(t), f_{l,z}(t)]^\top \in \mathbb{R}^3$ represents the current focal point position. Critically, this local observation excludes information about other users $\{u_{k'} : k' \neq \pi(l)\}$ and other reflector segments $\{f_{l'} : l' \neq l\}$, reducing the observation space dimensionality from $\mathbb{R}^{3K+6L}$ of a joint MARL to $\mathbb{R}^9$ at the low level, a reduction factor of~$(K+2L)/3$.

The observation masking principle exploits the locality structure of the wireless optimization problem: under an effective allocation, each reflector segment's focal point primarily affects its assigned user's signal strength, with limited impact on other users. This assumption enables independent optimization while maintaining near-optimal system-wide performance.

\textbf{Action Space:} Each low-level controller adjusts its focal point position through continuous displacement actions:
\begin{equation}
a_{l,t} = [\Delta f_{l,x,t}, \Delta f_{l,y,t}, \Delta f_{l,z,t}]^\top \in \mathcal{A}_L \subset \mathbb{R}^3,
\label{eq:low_level_action}
\end{equation}
subject to the maximum displacement constraint:
\begin{equation}
\|a_{l,t}\|_\infty \leq \delta_{\max},
\label{eq:action_constraint}
\end{equation}
where $\delta_{\max}$ defines the maximum allowable displacement per time step, typically set based on mechanical actuation capabilities. The focal point evolves according to the dynamics:
\begin{equation}
f_l(t+1) = f_l(t) + a_{l,t},
\label{eq:focal_point_dynamics}
\end{equation}
with the constraint that $f_l(t+1) \in \mathcal{F}_l$, where $\mathcal{F}_l$ is the feasible focal point region defined in the previous section.

\textbf{Subtask Decomposition and Local Value Function:} The hierarchical architecture decomposes the global value function into subtask-specific components, with each low-level controller learning a local Q-function $Q_{L,l}: \mathcal{S}_{L,l} \times \mathcal{A}_L \rightarrow \mathbb{R}$.
\begin{equation}
Q_{L,l}(s_{L,l}, a_l) = \mathbb{E}\left[\sum_{t=0}^{\infty} \gamma^t R_l(s_{L,l}(t), a_l(t))\right],
\label{eq:low_level_Q}
\end{equation}
where $R_l(s_{L,l}, a_l(t)) = P_{r,\pi(l)}(u_{\pi(l)}, f_l)$ represents the RSSI for the assigned user, as defined in Equation~\eqref{eq:received_power}. The subtask decomposition leverages the assumption that, under optimal allocation and with appropriate focal point control, the received power at user $k$ is primarily determined by the focal point of its assigned reflector segment. We assume that cross-reflector interference terms are negligible when users are spatially separated, and allocation is geometrically appropriate. This decomposition enables parallel learning of $L$ independent focal point policies while maintaining near-optimal system-wide coordination.

\textbf{Low-Level Optimization Objective:} Each low-level agent seeks a policy $\pi_{L,l}: \mathcal{S}_{L,l} \rightarrow \mathcal{A}_L$ that maximizes the expected RSSI for its assigned user:
\begin{equation}
\pi_{L,l}^* = \mathop{\arg\max}_{\pi_{L,l}} \mathop{\mathbb{E}}_{s_{L,l}, a_l \sim \pi_{L,l}}\left[\sum_{t=0}^{\infty} \gamma^t P_{r,\pi(l)}(u_k(t), f_{l}(t))\right],
\label{eq:low_level_objective}
\end{equation}
subject to the focal point dynamics in Equation~\eqref{eq:focal_point_dynamics}, feasibility constraints $f_l(t) \in \mathcal{F}_l$, and action bounds in Equation~\eqref{eq:action_constraint}.

\subsubsection{Unified Hierarchical Optimization Problem}

The complete reflective surface optimization problem integrates the high-level allocation and low-level focal point control subproblems through a shared system-wide performance objective. The unified formulation is expressed as a hierarchical constrained optimization:

\begin{maxi!}|s|
    {b, \{f_l\}_{l=1}^L}{\mathbb{E}\left[\sum_{t=0}^{\infty} \gamma^t R_{\text{sys}}(s(t), b(t))\right] \label{eq:unified_optimization}}
    {}{}
    \addConstraint{b(t) \in \mathcal{B}, \quad \forall t \in \{0, T, 2T, \ldots\} \label{eq:const_allocation}}
    \addConstraint{f_l(t+1) = f_l(t) + a_{l,t}, \quad \forall l, t \label{eq:const_dynamics}}
    \addConstraint{f_l(t) \in \mathcal{F}_l, \quad \forall l, t \label{eq:const_feasible}},
    \addConstraint{\|a_{l,t}\|_\infty \leq \delta_{\max}, \quad \forall l, t \label{eq:const_action}}
    \addConstraint{\theta_{i,j}(f_l(t)) \in [\theta_{\min}, \theta_{\max}], \quad \forall (i,j) \in \mathcal{S}_l, \forall l, \forall t \label{eq:const_theta}}
    \addConstraint{\phi_{i,j}(f_l(t)) \in [\phi_{\min}, \phi_{\max}], \quad \forall (i,j) \in \mathcal{S}_l, \forall l, \forall t \label{eq:const_phi}},
\end{maxi!}
where the complete system state $s(t)$ coincides with the high-level state: $s(t)=s_H(t)$; We use $s_H(t)$ when discussing the high-level controller specifically and $s(t)$ when referring to the complete system state. 
We emphasize that the physical constraints specified in (\ref{eq:const_feasible})--(\ref{eq:const_phi}) are rigorously enforced by the simulation environment as hard limits. At each time step, focal point positions are strictly clipped to the feasible region $\mathcal{F}_l$, while tile orientations are mechanically restricted to the servo's operational range. This inherent bounding of the action space ensures all agent decisions remain physically valid, thereby removing the need for auxiliary penalty terms or Lagrange multipliers typically required in Constrained MDP formulations.

This unified formulation exhibits three critical properties that enable effective hierarchical learning:

\textbf{Coupling Structure:} The system reward factorizes across users but couples through the allocation mapping $b$, creating incentives for the high-level controller to assign users to reflector segments that can effectively serve them, while low-level controllers optimize focal points for their assigned users.

\textbf{Temporal Decoupling:} The temporal abstraction at the high level ($T$ time steps) creates a separation of time scales between high-level allocation and low-level execution, allowing low-level policies to partially converge before reassignment occurs. This temporal structure is essential for stable hierarchical learning.

\textbf{Constraint Integration:} Physical constraints (tile orientation limits, focal point feasibility) are naturally enforced through the focal point abstraction and feasible region $\mathcal{F}_l$, avoiding the need for explicit constraint satisfaction at the low level during exploration.

\begin{figure}[!t]
    \centering
    \captionsetup{justification=centering}
    \includegraphics[width=1.0\linewidth]{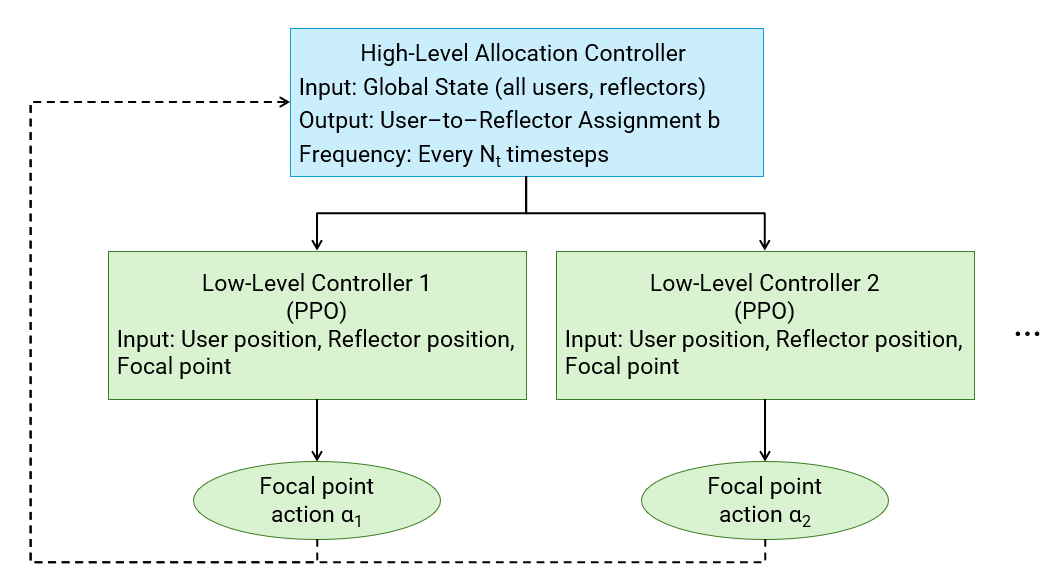}
    \caption{{Hierarchical Multi-Agent Reinforcement Learning Architecture. The high-level controller performs high-level user-to-reflector allocation using global system state at extended temporal intervals ($T$ timesteps), while low-level controllers execute low-level focal point optimization using masked local observations at every timestep. The framework employs Hierarchical Multi-Agent Reinforcement Learning (HMARL), enabling global state access during training while maintaining practical deployment scalability through local observation-based execution. Dashed arrows represent aggregated reward feedback from low-level controllers to the high-level coordinator.}}
    \label{Figure:hierarchical_architecture}
\end{figure}

\section{Hierarchical Multi-Agent Reinforcement Learning Framework}
\label{sec:marl_framework}

This section details the hierarchical learning framework enabling efficient CSI-free optimization, covering the coordination mechanisms within a Centralized Training with Decentralized Execution (CTDE) architecture, the Multi-Agent Proximal Policy Optimization (MAPPO) algorithm, and the integration of a geometric compatibility matrix for learning acceleration.

\subsection{Hierarchical Coordination and CTDE Framework}
\label{subsec:hierarchical_coordination}

The hierarchical framework coordinates high-level allocation and low-level execution through structured information flow and shared performance objectives while leveraging CTDE to address multi-agent non-stationarity challenges. Fig.~\ref{Figure:hierarchical_architecture} illustrates the complete hierarchical MARL architecture, showing the information flow between high-level allocation and low-level execution levels.

\subsubsection{Information Flow and Reward Coupling}

The hierarchical coordination employs multi-directional information flow that couples learning across levels while maintaining computational scalability. During execution, the high-level controller observes global state $s_H(t)$ at discrete time steps $t \in \{0, T, 2T, \ldots\}$ and selects allocation $b(t)$ that determines user-to-reflector assignments. This allocation decision propagates to all low-level controllers, establishing the assignment mapping $\pi(l)$ that specifies which user each reflector segment serves. The high-level decision remains fixed for $T$ timesteps, providing a stable optimization horizon for low-level adaptation.

Each low-level controller $l$ operates autonomously at every timestep, observing only its masked local state $s_{L,l}(t) = \{u_{\pi(l)}(t), r_l, f_l(t)\}$ containing the position of its assigned user, its own reflector position, and its current focal point. This decentralized execution preserves scalability and enables parallel implementation across reflector segments without inter-agent communication. During training, system-wide performance $R_{\text{sys}}(s(t), b(t))$ aggregates received power contributions from all users, providing bottom-up feedback to guide high-level allocation learning.

\subsubsection{Centralized Training with Decentralized Execution}

The framework employs the CTDE paradigm for the low-level control to address multi-agent non-stationarity while maintaining practical deployment scalability. During centralized training, a global critic network $V^\pi(s_{\text{global}})$ accesses complete system state $s_{\text{global}} = \{u_1, \ldots, u_K, r_1, \ldots, r_L, f_1, \ldots, f_L\}$ encompassing all user positions, reflector positions, and focal point configurations. $s_{\text{global}}$ is equivalent to $s(t)$ but emphasizes training-time access. The global critic estimates the expected cumulative return:
\begin{equation}
V^\pi(s_{\text{global}}) = \mathbb{E}_\pi[G_t \mid S_{\text{global},t} = s_{\text{global}}],
\label{eq:global_value_function}
\end{equation}
where $G_t = \sum_{t=0}^{\infty} \gamma^t R_{\text{sys}}(t+t)$. This global value function enables advantage estimation for policy updates:
\begin{equation}
\text{Adv}(s_{\text{global}}, b) = Q^\pi(s_{\text{global}}, b) - V^\pi(s_{\text{global}}),
\label{eq:advantage_estimation}
\end{equation}
stabilizing learning by providing consistent value estimates despite the non-stationary environment induced by concurrent policy updates across multiple agents. Specifically, purely decentralized agents would perceive the evolving policies of peers as unpredictable environmental shifts, violating the standard Markov assumption. The global critic resolves this by conditioning on the joint state, effectively alleviate the non-stationarity of the learning target during the training phase. During training, Advantage uses global state for accurate credit assignment, but policies are conditioned only on local observations, ensuring decentralized execution. During execution, the global critic is discarded, and agents execute policies based solely on their prescribed local observations. The high-level policy $\pi_H(b|s_H)$ uses complete information to perform the user-reflector assignment. Low-level policies $\pi_{L,l}(a_l|s_{L,l})$ use only their assigned user's position, own reflector position, and current focal point. This decentralized execution eliminates communication overhead, enables parallel computation across reflector segments, and preserves privacy.

\subsubsection{Joint Learning and Policy Modularity}

Both hierarchical levels learn concurrently through shared experiences, with high-level allocation decisions creating the context for low-level focal point optimization. The temporal abstraction mechanism ensures learning stability by separating high-level and low-level time scales: the high-level allocation persists for $T$ timesteps while low-level controllers adapt their focal points. This temporal commitment prevents destructive policy oscillations that would arise from simultaneous high-frequency changes in both allocation and focal point configurations, reducing variance in advantage estimates and accelerating convergence.

% The hierarchical formulation provides substantial computational and learning efficiency advantages. The observation masking reduces state space dimensionality from $\mathbb{R}^{3K+6L}$ (required for full observability) to $\mathbb{R}^9$ per low-level agent, yielding a reduction factor of $(K+2L)/3$. For a representative system with $K=4$ users and $L=4$ segments, this corresponds to approximately $4\times$ dimensional reduction, directly translating to decreased neural network capacity requirements and improved sample efficiency.

\subsection{Multi-Agent Proximal Policy Optimization}
\label{subsec:mappo}

Both high-level and low-level controllers employ Multi-Agent Proximal Policy Optimization (MAPPO), which extends the single-agent PPO algorithm~\cite{schulman2017proximal} to multi-agent settings while addressing coordination and non-stationarity challenges~\cite{yu2022surprising}. For each agent $l$, the clipped surrogate objective incorporates global state information from the CTDE architecture:

\begin{equation}
\begin{aligned}
L^{CLIP}(\psi_l) = \mathbb{E}_t \Big[ \min\Big( r_t^l(\psi_l) \text{Adv}(s_{global,t}, a_t^l), \\
\text{clip}(r_t^l(\psi_l), 1{-}\epsilon, 1{+}\epsilon) \text{Adv}(s_{global,t}, a_t^l) \Big) \Big],
\end{aligned}
\end{equation}
where the probability ratio $r_t^l(\psi_l) = \frac{\pi_{\psi_l}(a_t^l|o_t^l)}{\pi_{\psi_{\text{old}},l}(a_t^l|o_t^l)}$, with $o_t^l$ is the partial observability from the complete state, quantifies the policy modification between updates. The clipping parameter $\epsilon$ (typically 0.1-0.2) constrains policy modifications to ensure training stability, preventing destructive updates that might disrupt learned coordination mechanisms.

The value function loss for the centralized critic is:

\begin{equation}
\mathcal{L}^{VF}(\xi) = \mathbb{E}_t\left[(V_\xi(\mathbf{s}_{\text{global},t}) - V_{\text{target},t})^2\right],
\end{equation}
where $V_{\text{target},t}$ denotes the target value calculated using trajectory information. The comprehensive MAPPO objective integrates policy and value function losses with entropy regularization:

\begin{equation}
\mathcal{L}(\psi_l, \xi) = \mathcal{L}^{\text{CLIP}}(\psi_l) - c_1\mathcal{L}^{VF}(\xi) + c_2\mathcal{H}_{\pi_{\psi_l}},
\end{equation}
where $\mathcal{H}_{\pi_{\psi_l}}$ denotes entropy regularization to promote exploration, and $c_1, c_2$ represent scaling coefficients.

\subsection{Compatibility Matrix for Learning Acceleration}
\label{subsec:compatibility_matrix}

To accelerate convergence during early training phases, we augment the high-level allocation controller with a domain-specific compatibility matrix $\mathbf{C} \in \mathbb{R}^{K \times L}$ that encodes prior geometric knowledge about user-reflector assignment quality. The matrix element $C_{kl}$ represents the expected signal propagation favorability when user $k$ is assigned to reflector $l$, quantified through geometric relationships:

\begin{equation}
C_{kl} = \exp\left(-\frac{\|\mathbf{u}_k - \mathbf{r}_l\|}{d_0}\right) \cdot \cos(\theta_{kl}) ,
\end{equation}
where $\|\mathbf{u}_k - \mathbf{r}_l\|$ denotes the Euclidean distance between user $k$ and reflector $l$, $d_0$ is a normalization constant, and $\theta_{kl}$ represents the AP-reflector-user reflection angle. High compatibility scores indicate geometric configurations where the reflector can efficiently redirect signal power toward the user with minimal path loss and favorable incident angles.

We integrate the compatibility matrix into the high-level allocation policy through a weighted formulation:

\begin{equation}
\pi_H(b|\mathbf{s}_H; \phi) \propto \left(Q_H(\mathbf{s}_H, b) + \alpha(t) \sum_{k=1}^K C_{k,b_k}\right) ,
\end{equation}
where $Q_H(\mathbf{s}_H, b)$ is the learned Q-value for allocation $b$ in state $\mathbf{s}_H$, and $\alpha(t)$ is a coefficient that reduces the influence of prior knowledge as training progresses. Specifically, we employ:

\begin{equation}
\alpha(t) =
\begin{cases}
    1 & \text{t $<$ threshold}\\
    0 & \text{t $\geq$ threshold}
\end{cases}
\end{equation}

This design philosophy differs from conventional imitation learning or reward shaping approaches in that the compatibility matrix serves as a critical inductive bias for achieving high performance in sparse reward environments. As demonstrated in Section~\ref{sec:results_and_discussion}.\ref{subsec:training}, the matrix not only accelerates early convergence by approximately 200-300 episodes in 2-user scenarios, but also enables the hierarchical allocation mechanism to achieve superior final performance (37-28\% improvement) compared to learning without domain-specific guidance. This demonstrates that while the framework \textit{can} learn assignment strategies through exploration alone, the exponentially large combinatorial action space makes discovering near-optimal allocations impractical within typical training horizons without geometric priors.

\section{Results and Discussion}
\label{sec:results_and_discussion}

\paragraph*{Section Overview.}
This section presents a structured evaluation of the proposed hierarchical MARL framework across realism, learning behavior, deployment performance, and robustness. We begin by describing the experimental setup and simulation framework, including the mmWave propagation environment, system configuration, and baseline methods. We then analyze training convergence to isolate the impact of hierarchical decomposition and the compatibility matrix. Deployment-time RSSI performance is subsequently evaluated under user mobility, followed by scalability analysis with increasing user density. Hardware–performance trade-offs are examined through reflector aperture size variation, and robustness is assessed via reward function sensitivity and user localization error analysis. Together, these results provide a comprehensive assessment of learning efficiency, deployment viability, and practical system design trade-offs.

\begin{figure}[!t]
    \centering
    \captionsetup{justification=centering}
    \includegraphics[width=1.0\linewidth]{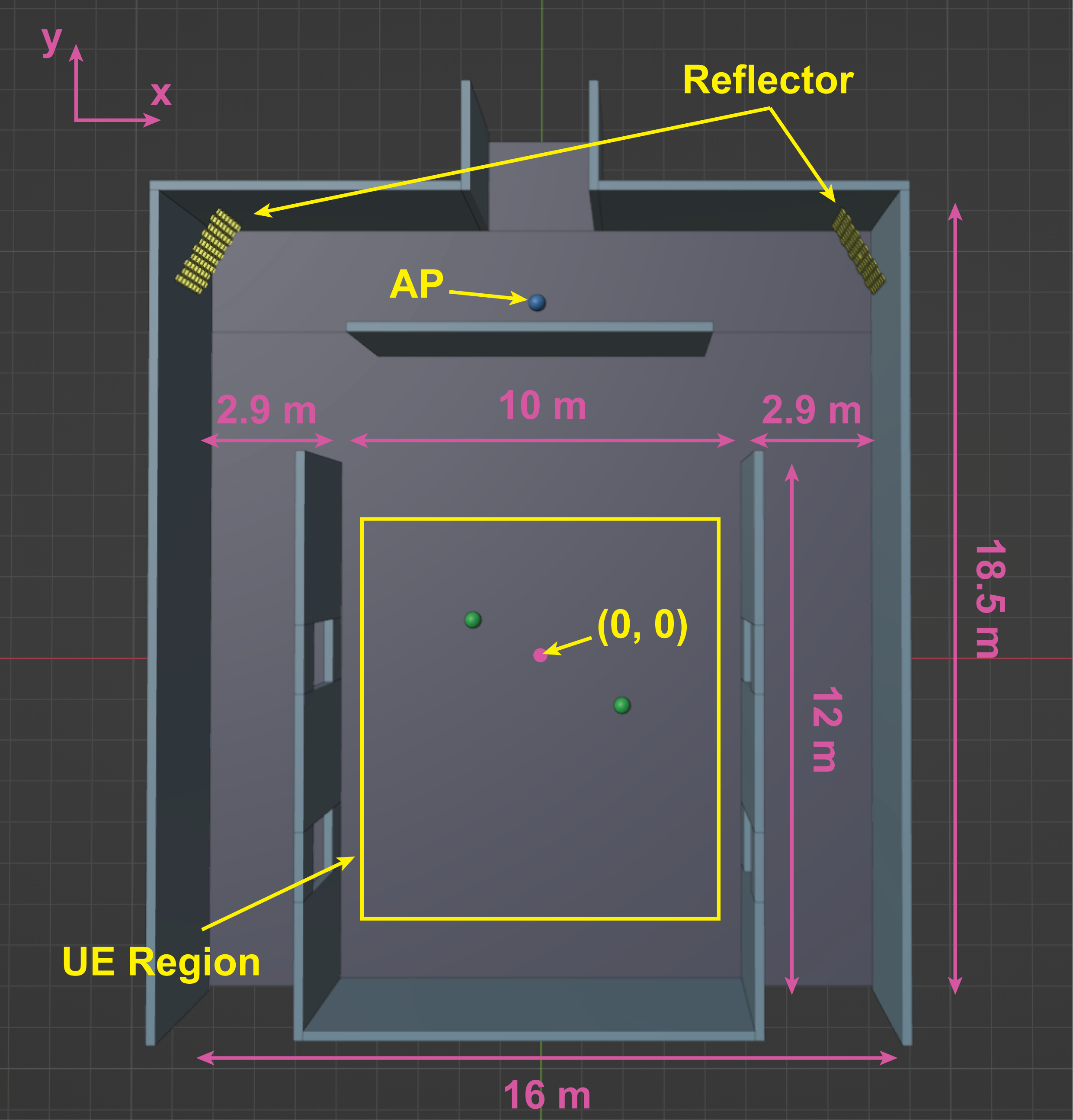}
    \caption{{Experimental setup of a conference room for simulation. The AP is depicted in blue, while the users are shown in green. The users are within the UE region.}}
    \label{Figure:experimental_setup_classroom}
\end{figure}

\subsection{Experimental Setup and Simulation Framework}
\label{subsec:experimental_setup}

The performance evaluation of the proposed HMARL framework is conducted in a realistic indoor millimeter-wave environment using a high-fidelity simulation platform that integrates NVIDIA Sionna's deterministic ray-tracing engine \cite{sionna} with Blender's three-dimensional modeling capabilities \cite{blender}. This integrated computational framework enables accurate electromagnetic wave propagation modeling while incorporating material-specific properties defined according to ITU-R Recommendation P.2040-1 \cite{rec_p2040}, ensuring practical relevance and deployment fidelity.

\subsubsection{Simulation Environment and Propagation Modeling}

The experimental testbed comprises a conference room scenario where an access point (AP) is positioned outside the room to serve multiple coverage areas, creating significant propagation challenges due to structural wall attenuation. Two mechanically reconfigurable metallic reflector arrays are high-level deployed at corner positions to redirect millimeter-wave signals toward user equipment within the conference room. The simulation environment employs deterministic ray-tracing to capture multi-path propagation phenomena including reflections, diffractions, and scattering effects characteristic of indoor millimeter-wave channels.

Material properties for all surfaces, including walls (concrete with relative permittivity $\epsilon_r = 5.31$, conductivity $\sigma = 0.0326$ S/m), ceilings (wood with $\epsilon_r = 2.89$, $\sigma = 0.0047$ S/m), and floors (marble with $\epsilon_r = 7.0$, $\sigma = 0.01$ S/m), are specified according to ITU-R P.2040-1 standards to ensure realistic propagation loss characteristics \cite{rec_p2040}. The reflector surfaces are modeled as metallic panels with near-perfect reflection coefficients ($\epsilon_r = 1$, high conductivity), enabling accurate representation of specular reflection and signal redirection properties. Each reflector consists of independently adjustable hexagonal metallic tiles with mechanical servo-based control, providing wideband frequency operation without requiring sophisticated RF circuitry or electronic phase shifters.

\subsubsection{System Configuration and Training Parameters}

The carrier frequency is set to 60~GHz with a total transmit power of 5~dBm, representing mmWave communication systems operating with low transmitting power. Fig.~\ref{Figure:experimental_setup_classroom} illustrates the experimental setup of this work. User equipment positions are sampled uniformly within a designated coverage region measuring approximately 10~m $\times$ 10~m, with random repositioning at the start of each training episode to promote policy generalization across diverse spatial configurations. Reflector focal points are initialized around the center of the scenario, according to a three-dimensional Gaussian distribution characterized by mean vector $\boldsymbol{\mu} = [0.0, 0.0, 1.5]$~m and covariance matrix $\boldsymbol{\Sigma} = 2.5 \mathbf{I}$, where $\mathbf{I}$ denotes the identity matrix. This stochastic initialization strategy ensures comprehensive exploration of the reflector orientation space while maintaining physical feasibility constraints.

The MAPPO algorithm operates under a CTDE paradigm, where global system state information is accessible during training and at a centralized high-level controller but individual low-level agents execute policies based solely on local observations during deployment. The high-level controller consists of an attention layer with a layer normalization followed by 2 fully connected heads (128 neurons), one is for the Q-value and the other is the policy action. Each low-level agent's policy and value network consists of two fully connected layers with 256 neurons and ReLU activation functions, providing sufficient representational capacity for learning complex coordination behaviors. The Adam optimizer with a learning rate of $2.0 \times 10^{-4}$ ensures stable policy updates, while a discount factor of $\gamma = 0.985$ emphasizes long-term cumulative rewards. PPO-specific parameters include a clipping parameter $\epsilon = 0.2$ to constrain policy modifications and a generalized advantage estimation (GAE) lambda of 0.9 for advantage estimation. Training proceeds over 3,200 episodes for both 2-user and 4-user scenarios, with each batch of 200 samples processed for 40 epochs to maximize sample efficiency. Each episode lasts 100 time steps. Table~\ref{tab:system_params} summarizes the complete system parameters and training hyperparameters.

\subsubsection{Performance Metrics and Baseline Methods}

The primary evaluation metric is the RSSI, measured in dBm, which quantifies the signal power at each UE location. System-level performance is assessed through both individual user RSSI and aggregate metrics, including mean RSSI across all users. Additionally, training convergence is evaluated using episode-averaged cumulative reward.

To comprehensively evaluate the proposed hierarchical architecture, four distinct control strategies are compared:
\begin{itemize}
    \item \textbf{Allocator}: Implements the full hierarchical framework with a high-level user-to-reflector allocation controller augmented by a compatibility matrix encoding prior domain knowledge, combined with low-level PPO-based focal point optimization agents.

    \item \textbf{No\_compat}: A variant that removes the compatibility matrix while retaining the hierarchical allocation-execution structure, enabling assessment of the matrix's contribution to learning acceleration.

    \item \textbf{No\_allocator}: A baseline that employs a single centralized PPO agent that directly optimizes the average RSSI across all users without hierarchical decomposition, representing a conventional centralized optimization approach.

    \item \textbf{Random}: A baseline that assigns users to reflectors through uniform random selection while maintaining low-level PPO optimization for assigned users, establishing a lower-bound performance reference.
\end{itemize}

% To comprehensively evaluate the proposed hierarchical architecture, four distinct control strategies are compared. The \textit{Allocator} method implements the full hierarchical framework with a high-level user-to-reflector assignment controller augmented by a compatibility matrix encoding prior domain knowledge, combined with low-level PPO-based focal point optimization agents. The \textit{No\_compat} variant removes the compatibility matrix while retaining the hierarchical allocation-execution structure, enabling assessment of the matrix's contribution to learning acceleration. The \textit{No\_allocator} baseline employs a single centralized PPO agent that directly optimizes the average RSSI across all users without hierarchical decomposition, representing a conventional centralized optimization approach. Finally, the \textit{Random} baseline assigns users to reflectors through uniform random selection while maintaining low-level PPO optimization for assigned users, establishing a lower-bound performance reference.

The selection of these baselines is designed to isolate and verify the distinct contributions of the proposed framework. Specifically, \textbf{No\_allocator} serves as the architectural benchmark, validating the hypothesis that hierarchical decomposition is superior to standard centralized learning for high-dimensional control problems. \textbf{No\_compat} functions as an ablation study, isolating the specific performance gain configurable to the geometric inductive bias. Finally, \textbf{Random} establishes the performance floor, confirming that the user-to-reflector assignment is a non-trivial combinatorial problem that requires intelligent optimization rather than stochastic selection.

\subsubsection{Evaluation Protocol}

Post-training evaluation is conducted over 300 timesteps with fixed trained policies to assess deployment performance and RSSI stability. To assess robustness, realistic mobility is simulated via 30 cm user displacements every two steps under varying reflector configurations. All experiments are executed with deterministic ray-tracing to ensure reproducibility, capturing multi-path propagation phenomena characteristic of indoor millimeter-wave channels.

\begin{table}[htbp]
\caption{System parameters and training hyperparameters}
\label{tab:system_params}
\centering
\begin{tabular}{ll}
\hline
\textbf{Parameter} & \textbf{Value} \\
\hline
\multicolumn{2}{l}{\textbf{System Configuration}} \\
Carrier frequency \& transmit power & 60 GHz, 5 dBm \\
User scenarios \& reflector arrays & 2, 4 users; 2 arrays \\
Coverage area & 10 m $\times$ 10 m \\
\\ 
\multicolumn{2}{l}{\textbf{Neural Network \& Optimization}} \\
High-level layer \& activation & attention + 128 neurons, ReLU \\
Low-level hidden layers \& activation & 256 neurons, ReLU \\
Optimizer \& learning rate & Adam, $2.0 \times 10^{-4}$ \\
\\ 
\multicolumn{2}{l}{\textbf{PPO Configuration}} \\
Discount factor $\gamma$, clip $\epsilon$, GAE $\lambda$ & 0.985, 0.2, 0.9 \\
Entropy \& value loss coefficients & $1.0 \times 10^{-4}$, 1.0 \\
Batch size \& training epochs & 200, 40 \\
\\ 
\multicolumn{2}{l}{\textbf{Training \& Evaluation}} \\
Training episodes \& eval timesteps & 3,200, 300 \\
Focal point initialization $\boldsymbol{\mu}$, $\boldsymbol{\Sigma}$ & [0.0, 0.0, 1.5] m, $2.5 \mathbf{I}$ \\
\hline
\end{tabular}
\end{table}

\subsection{Training Performance and Convergence Analysis}
\label{subsec:training}

\begin{figure}[!t]
    \centering
    \captionsetup{justification=centering}
    \includegraphics[width=1.0\linewidth]{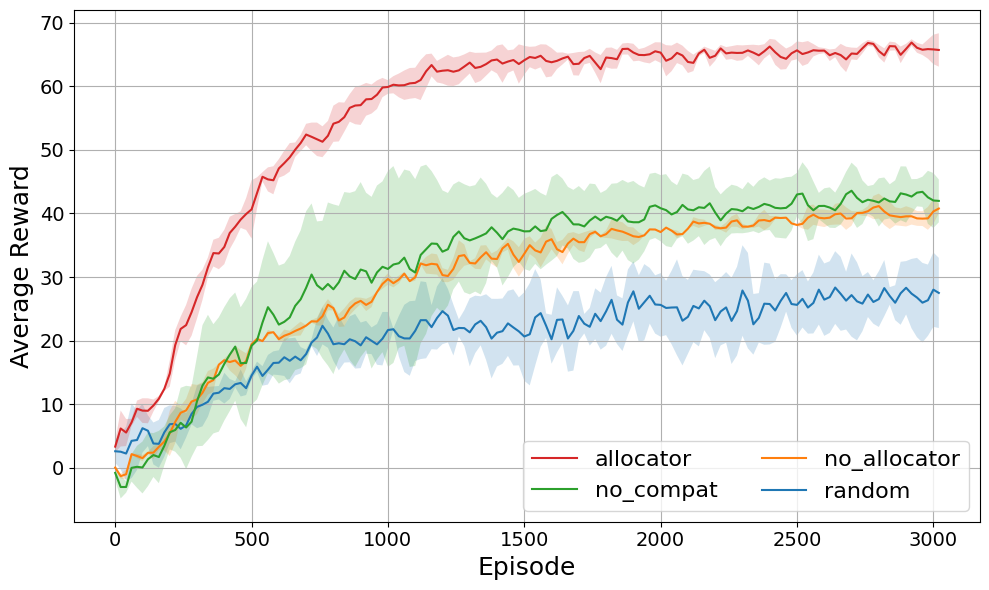}
    \caption{{Training performance of the two-user scenario: episode-averaged reward for two-user scenario across four methods over 3,200 episodes.}}
    \label{Figure:training_2ue}
\end{figure}

\begin{figure}[!t]
    \centering
    \captionsetup{justification=centering}
    \includegraphics[width=1.0\linewidth]{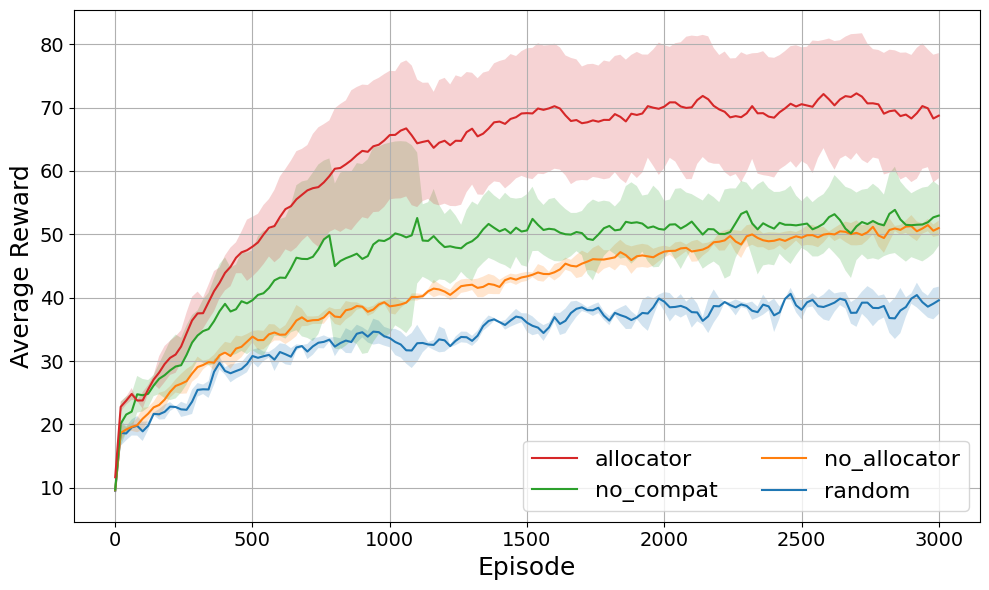}
    \caption{{Training performance of the four-user scenario: episode-averaged reward for four-user scenario across four methods over 3,200 episodes.}}
    \label{Figure:training_4ue}
\end{figure}

The training performance of the hierarchical MARL framework is evaluated across 2-user (2UE) and 4-user (4UE) configurations. Fig.~\ref{Figure:training_2ue} and Fig.~\ref{Figure:training_4ue} present episode-averaged reward convergence over 3,200 training episodes, comparing four control strategies.

The \textit{Allocator} method achieves superior convergence in both scenarios, reaching average rewards of approximately 67 (2UE) and 69 cumulative reward (4UE) after convergence. This demonstrates rapid initial learning within 500 episodes followed by stable improvement, with consistent behavior regardless of user density scaling.

\textbf{Impact of Compatibility Matrix:} The performance difference between \textit{Allocator} and \textit{No\_compat} reveals the critical contribution of domain knowledge encoding. In the 2UE scenario (Fig.~\ref{Figure:training_2ue}), \textit{Allocator} achieves a final reward of approximately 67 cumulative reward, while \textit{No\_compat} converges to only 42 cumulative reward, representing a 37\% performance gap that persists throughout training. The 4UE scenario (Fig.~\ref{Figure:training_4ue}) exhibits similar behavior (69 vs. 50, 28\% gap). This substantial advantage demonstrates that the compatibility matrix provides essential inductive bias guiding the allocation controller toward geometrically favorable assignments. The exponentially large discrete action space ($K^L$ combinations) creates a sparse reward landscape where effective strategies require prohibitively extensive exploration without domain-specific guidance. While the compatibility matrix requires geometric knowledge readily available through localization infrastructure, the substantial performance improvement (37\% in 2UE, 28\% in 4UE) justifies the minimal overhead compared to CSI estimation in traditional RIS systems.

\textbf{Hierarchical vs. Centralized Learning:} Hierarchical methods (\textit{Allocator} and \textit{No\_compat}) outperform the centralized \textit{No\_allocator} baseline. In the 2UE scenario, hierarchical approaches achieve rewards of 67 versus 40 (67\% improvement), while in 4UE they reach 69 compared to 50 (38\% improvement) compared to centralized approaches. The centralized approach struggles due to increased observation dimensionality and credit assignment complexity when a single agent must simultaneously optimize focal points for multiple reflectors serving multiple users. The hierarchical decomposition distributes decision-making across specialized controllers operating at different temporal abstractions, enabling more efficient exploration and coordination.

\textbf{Baseline Performance:} The \textit{Random} baseline employs uniform random user-reflector assignment with low-level PPO optimization, achieving the poorest performance: approximately 27 (2UE) and 40 (4UE). This underperformance, even compared to centralized optimization, highlights the necessity of learned assignment strategies rather than naive heuristics for effective multi-reflector coordination.

\textbf{Scalability Across User Densities:} Comparing 2UE and 4UE scenarios reveals that the hierarchical framework scales effectively with increasing user density. While the final mean convergence performance remains comparable (69 for 2UE vs. 67 for 4UE), the 4-user scenario exhibits notably higher reward variance, as indicated by the wider shaded regions in Fig.~\ref{Figure:training_2ue} compared to Fig.~\ref{Figure:training_4ue}. This increased variance reflects the increased difficulty of simultaneously coordinating reflector assignments in a denser environment with a significantly larger combinatorial action space. Nevertheless, the system successfully converges to a high-performance policy, demonstrating that the hierarchical architecture maintains learning efficiency despite the increased complexity.

\subsection{Deployment Performance Evaluation and Architectural Comparison}
\label{subsec:deployment_performance}

\begin{figure}[!t]
    \centering
    \captionsetup{justification=centering}
    \includegraphics[width=1.0\linewidth]{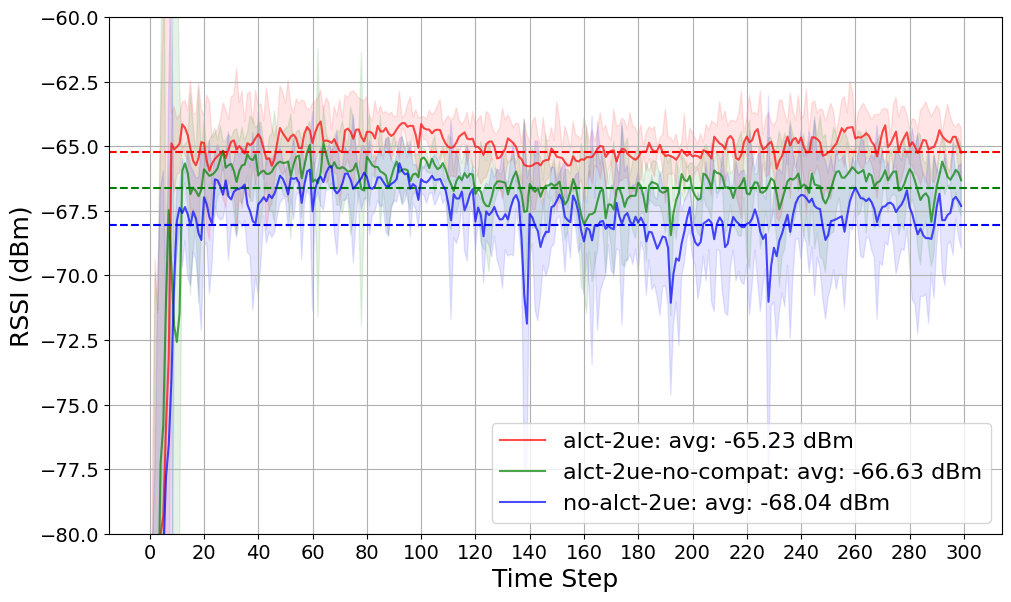}
    \caption{{Deployment RSSI Performance for Two-User Configuration: Hierarchical MARL with and without Compatibility Matrix, Centralized PPO, and Random Assignment. Solid lines represent mean RSSI, and shaded regions denote empirical standard deviation.}}
    \label{Figure:eval_2ue_methods}
\end{figure}

\begin{figure}[!t]
    \centering
    \captionsetup{justification=centering}
    \includegraphics[width=1.0\linewidth]{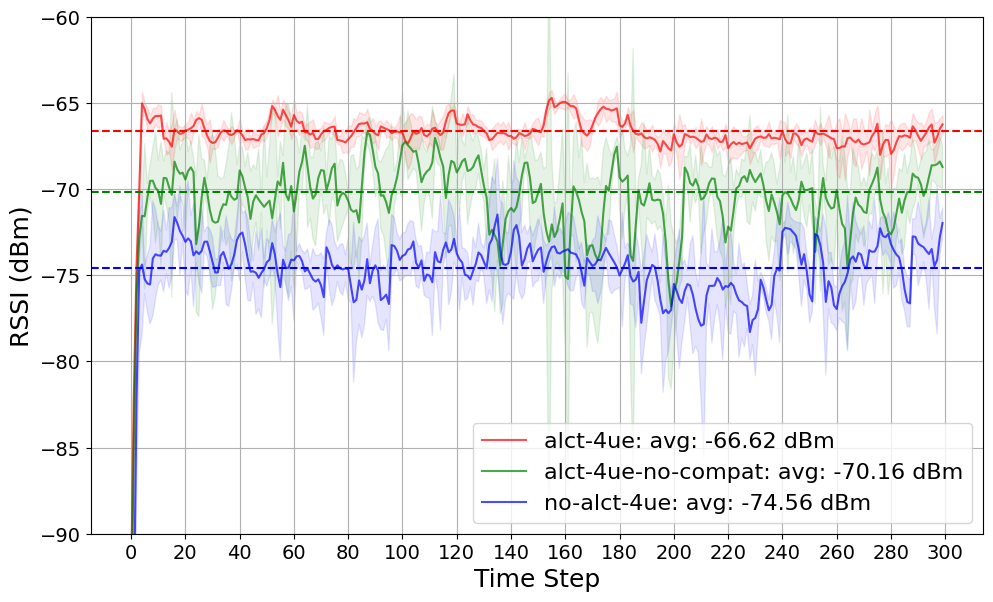}
    \caption{{Deployment RSSI Performance for Four-User Configuration: Hierarchical MARL with and without Compatibility Matrix, Centralized PPO, and Random Assignment. Solid lines represent mean RSSI, and shaded regions denote empirical standard deviation.}}
    \label{Figure:eval_4ue_methods}
\end{figure}

Post-training evaluation validates learned policy generalization and practical deployment performance. Fig.~\ref{Figure:eval_2ue_methods} and Fig.~\ref{Figure:eval_4ue_methods} present RSSI performance during 300-timestep evaluation with user movement with a velocity of $1$~m/s.

\textbf{Hierarchical vs. Centralized Optimization:} The hierarchical architecture achieves substantial RSSI improvements over centralized optimization. For 2-user configuration (Fig.~\ref{Figure:eval_2ue_methods}), \textit{Allocator} achieves $-65.23$ dBm compared to $-68.04$ dBm for centralized \textit{No\_allocator} (2.81 dB improvement). This advantage amplifies significantly in the 4-user scenario (Fig.~\ref{Figure:eval_4ue_methods}): $-66.62$ dBm versus $-74.56$ dBm (7.94 dB gain).

From the reinforcement learning perspective, hierarchical decomposition reduces both problem dimensionality and credit assignment complexity. The centralized agent must simultaneously optimize 3$L$ focal point parameters, resulting in action space growth and ambiguous reward attribution across reflector-user interactions. The hierarchical approach decouples this into high-level discrete allocation decisions with low-level continuous optimization maintaining fixed dimensionality per agent.

From the wireless communication perspective, the hierarchical framework enables task-specific optimization aligned with physical channel characteristics. The high-level allocator captures macro-scale geometric relationships and user-reflector compatibility, while low-level agents refine configurations based on fine-grained signal propagation dynamics. 
% This specialization exploits multi-user diversity, assigning complementary users to different reflectors reduces interference and signal saturation in resource-constrained scenarios.

\textbf{Scalability of Hierarchical Decomposition:} The performance gap between hierarchical and centralized methods widens with system complexity. The 4-user scenario exhibits a larger gap (7.94 dB) than the 2-user scenario (2.81 dB), demonstrating that centralized optimization suffers disproportionately as user density increases. Conversely, the hierarchical approach maintains consistent improvements, validating scalability to larger multi-user systems—crucial for mmWave networks with the dynamic user number.

\textbf{Compatibility Matrix Deployment Validation:} The \textit{Allocator} method consistently outperforms \textit{No\_compat}, achieving a $1.4$ dB gain in 2-user scenarios ($-65.23$ dBm vs. $-66.63$ dBm) and a $3.54$ dB gain in 4-user scenarios ($-66.62$ dBm vs. $-70.16$ dBm). These results confirm that the compatibility matrix effectively identifies and sustains optimal signal assignments. Crucially, this advantage persists because the policy retains the structured exploration patterns learned during training, proving that the method provides lasting allocation quality rather than just temporary training assistance.

\textbf{Stability and Variance Analysis:} Shaded regions in Fig.~\ref{Figure:eval_2ue_methods} and Fig.~\ref{Figure:eval_4ue_methods} represent empirical standard deviation. Hierarchical methods exhibit lower variance compared to centralized, particularly in 4-user scenarios. Stable RSSI performance reflects consistent assignment and optimization decisions despite dynamic positions. The centralized baseline shows larger fluctuations, degrading Quality of Service (QoS) guarantees.

\subsection{Scalability Analysis: User Density and System Efficiency Trade-off}

\begin{figure}[!t]
    \centering
    \captionsetup{justification=centering}
    \includegraphics[width=1.0\linewidth]{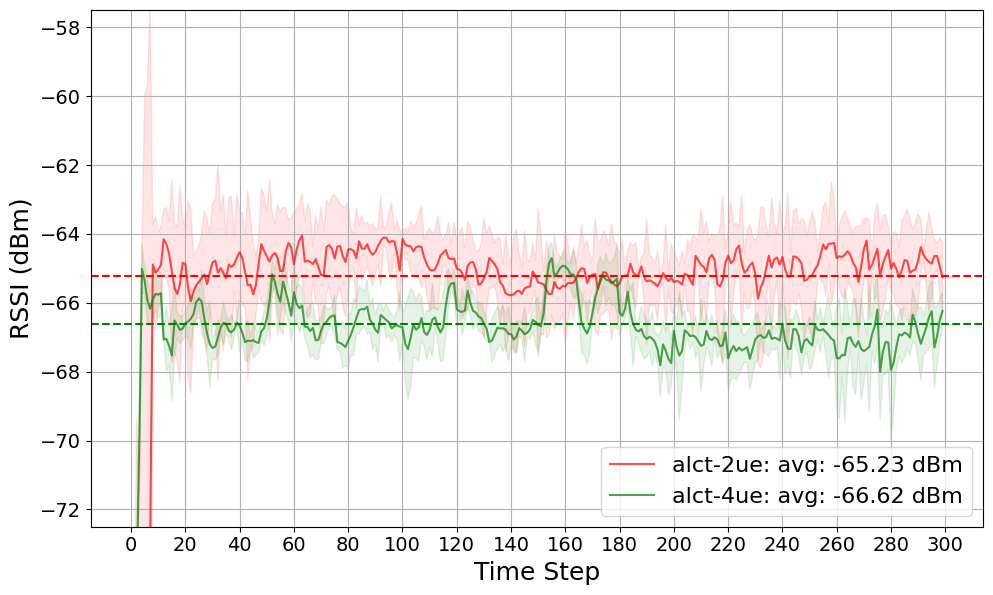}
    \caption{{Mean RSSI per User: Comparison of 2-User and 4-User Deployment Scenarios. Dashed lines indicate mean performance over 300 timesteps; shaded regions denote empirical standard deviation.}}
    \label{Figure:comparison_2ue_4ue_mean_power}
\end{figure}

\begin{figure}[!t]
    \centering
    \captionsetup{justification=centering}
    \includegraphics[width=1.0\linewidth]{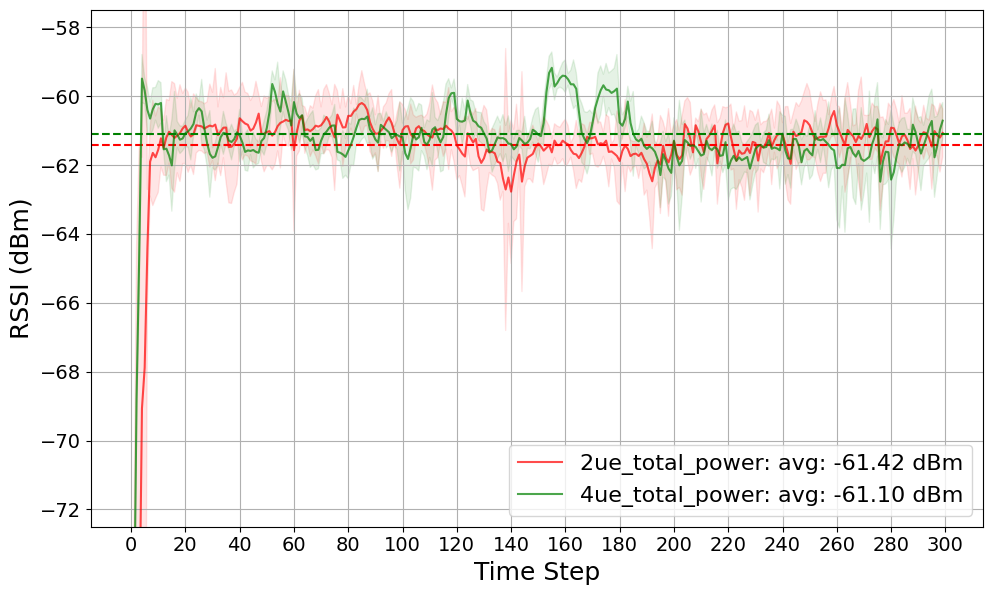}
    \caption{{Total System RSSI: Comparison of 2-User and 4-User Deployment Scenarios. Dashed lines indicate mean total performance over 300 timesteps; shaded regions denote empirical standard deviation.}}
    \label{Figure:comparison_2ue_4ue_total_power}
\end{figure}

To evaluate the scalability of the proposed HMARL framework, we analyze system performance under varying user densities. Fig.~\ref{Figure:comparison_2ue_4ue_mean_power} and Fig.~\ref{Figure:comparison_2ue_4ue_total_power} illustrate the mean RSSI per User and total system RSSI, respectively, for 2-user and 4-user configurations. These metrics shows the trade-off between individual Quality of Service (QoS) and aggregate system efficiency.

As demonstrated in Fig.~\ref{Figure:comparison_2ue_4ue_mean_power}, doubling the user density from 2 to 4 results in a per-user RSSI degradation of 1.39 dB (declining from $-65.23$ dBm to $-66.62$ dBm). Theoretical naive scaling, where fixed resources are divided equally among double the users, would necessitate a 3 dB reduction (50\% power loss). The observed degradation of only 1.39 dB indicates that the hierarchical allocator effectively mitigates power loss/interference, maintaining QoS standards better than a uniform resource split.

Fig.~\ref{Figure:comparison_2ue_4ue_total_power} reveals that the aggregate received power remains highly stable, with a negligible difference of 0.32 dB between the 2-user ($-61.42$ dBm) and 4-user ($-61.10$ dBm) scenarios. This consistency demonstrates that the algorithm maintains high efficacy in guiding signals toward assigned users, regardless of user density. Rather than suffering from saturation or interference losses as complexity grows, the hierarchical controller successfully identifies spatially complementary configurations, ensuring that the reflective surface area remains fully utilized to maximize total system throughput.

These results verify the framework's robustness in dynamic deployment scenarios. The system exhibits near-linear scalability, where the addition of users results in minimal per-user degradation (1.39 dB) while maintaining comparable total power levels. This confirms that the HMARL approach effectively exploits multi-user diversity to sustain efficient signal redirection.

\subsection{Impact of Reflector Aperture Size on System Performance}

\begin{figure}[!t]
    \centering
    \captionsetup{justification=centering}
    \includegraphics[width=1.0\linewidth]{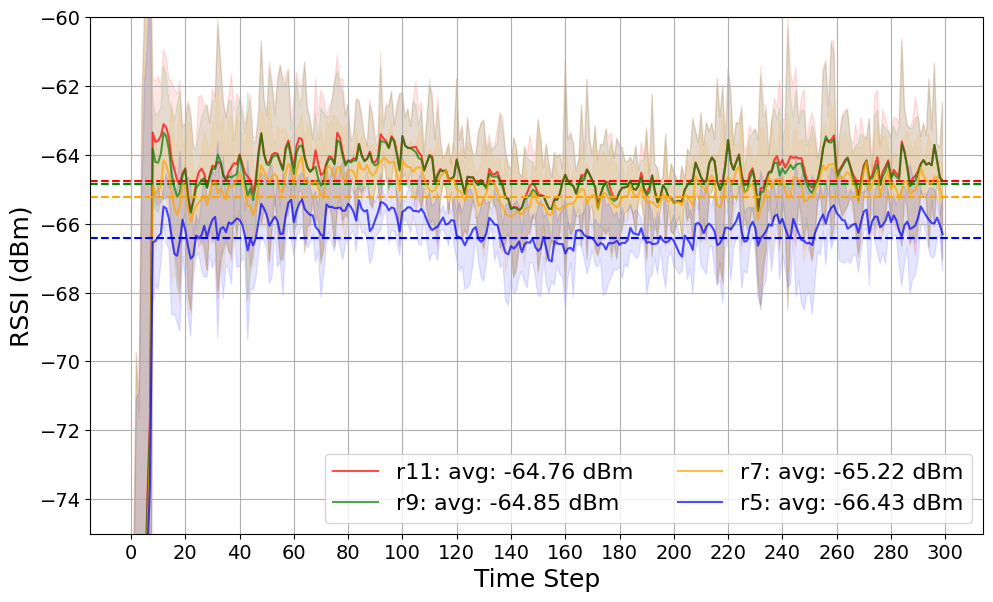}
    \caption{{RSSI Performance Across Reflector Aperture Configurations: 5-Row (45 tiles), 7-Row (63 tiles), 9-Row (81 tiles), and 11-Row (99 tiles) Implementations. Solid lines represent mean RSSI over 300 evaluation timesteps; shaded regions denote empirical standard deviation. Dashed horizontal lines indicate mean performance for each configuration.}}
    \label{Figure:eval_2ue_reflector_size}
\end{figure}

Optimizing the hardware configuration is essential for establishing a viable trade-off between deployment cost and system performance. To characterize this relationship, we evaluate the system's robustness across varying reflector aperture sizes. Fig.~\ref{Figure:eval_2ue_reflector_size} illustrates the RSSI performance for four distinct configurations, ranging from a compact 45-tile array (r5) to a large-scale 99-tile array (r11).

The evaluation reveals a monotonic relationship between aperture size and received signal strength, though with diminishing returns. The baseline r5 configuration (45 tiles) yields a mean RSSI of $-66.43$ dBm. Expanding the aperture to r7 (63 tiles) provides the most significant marginal gain of $1.21$ dB, resulting in $-65.22$ dBm. Subsequent increases yield progressively smaller improvements: the r9 configuration (81 tiles) adds $0.37$ dB (reaching $-64.85$ dBm), while the r11 configuration (99 tiles) adds a negligible $0.09$ dB (reaching $-64.76$ dBm).

Despite a $2.2\times$ increase in the number of reflective elements from r5 to r11, the total performance improvement is limited to $1.67$ dB. This saturation indicates that beyond a specific threshold, additional tiles contribute minimally to signal concentration due to the geometric constraints of the fixed user coverage area. Consequently, the r9 configuration emerges as the optimal design choice, capturing the majority of the achievable beam-focusing gain while minimizing hardware complexity compared to the r11 implementation.

\subsection{Reward Function Sensitivity Analysis}

\begin{figure}[!t]
    \centering
    \captionsetup{justification=centering}
    \includegraphics[width=1.0\linewidth]{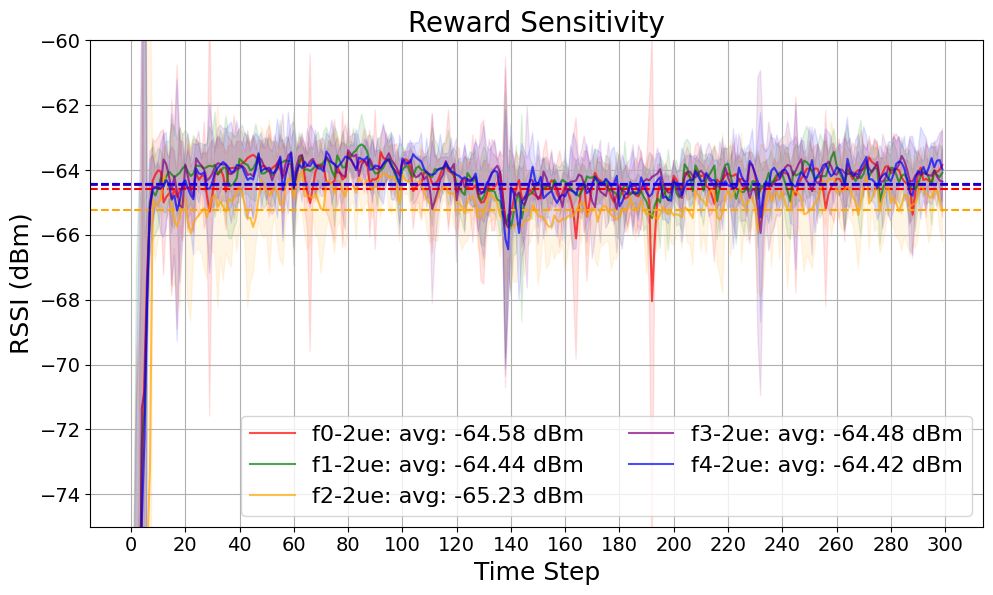}
    \caption{{Reward function sensitivity analysis: path-loss compensated reward signals. Path-Loss exponents $f_n \in \{f_0, f_1, f_2, f_3, f_4\}$. Solid lines represent mean RSSI over 300 evaluation timesteps; shaded regions denote empirical standard deviation.}}
    \label{Figure:eval_2ue_rew}
\end{figure}

To validate the robustness of the hierarchical framework to reward function design choices, we evaluate performance across five path-loss compensated reward formulations. The reward signal is constructed as $R = \text{RSSI} \times d^{-n}$, where $d$ is the propagation distance and $n \in \{0, 1, 2, 3, 4\}$ represents the path-loss exponent. This parameterization enables assessment of how different distance-normalization factors affect learning and deployment performance. Fig.~\ref{Figure:eval_2ue_rew} demonstrates that all five reward formulations converge to a narrow performance band spanning $-64.42$ to $-65.23$ dBm, representing only $0.81$ dBm stadard deviation across the tested r7 (63 tiles) configurations. This consistent performance confirms that the HMARL with MAPPO approach effectively mitigates the sensitivity issues common in reinforcement learning, maintaining robust performance without requiring meticulous reward engineering or specific distance scaling.

\subsection{Robustness to User Position Localization Errors}

\begin{figure}[!t]
    \centering
    \captionsetup{justification=centering}
    \includegraphics[width=1.0\linewidth]{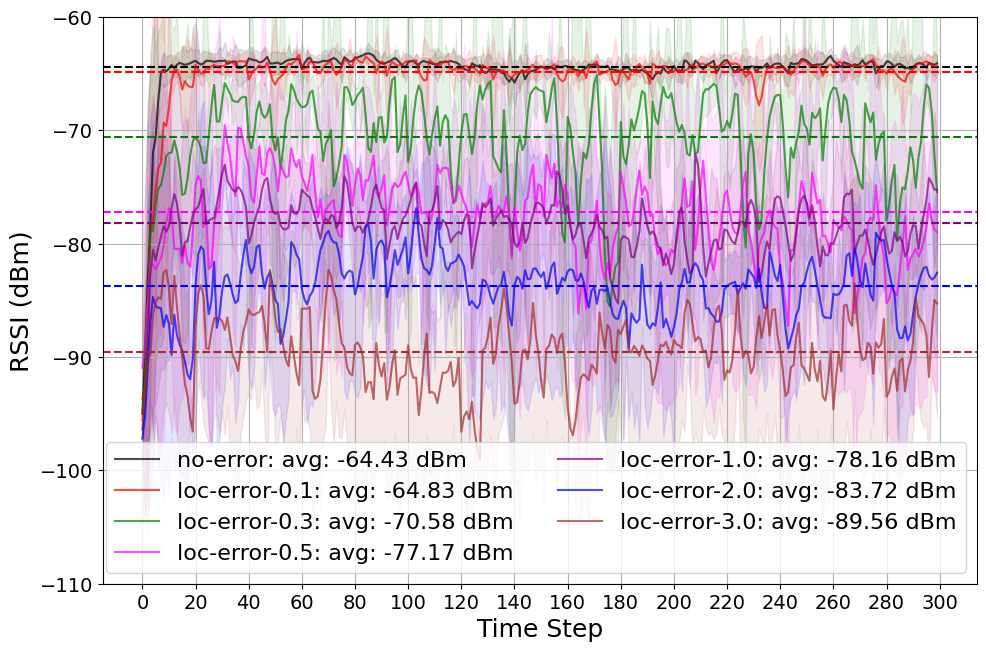}
    \caption{RSSI Performance Under User Position Localization Errors: 
    Performance across localization error levels 
    ($\sigma_{\text{error}} \in [0.1, 3.0]$ m), 
    evaluated with dynamic user repositioning. Solid lines represent mean RSSI 
    over 300 evaluation timesteps; shaded regions denote empirical standard 
    deviation. Results demonstrate graceful degradation up to 0.5 m error, 
    establishing practical localization accuracy requirements for reliable deployment.}
    \label{Figure:localization_error_robustness}
\end{figure}

Practical deployment of the hierarchical MARL framework depends on user localization information. Real-world localization systems encounter positioning errors due to hardware limitations and environmental multipath, which can degrade allocation quality if not properly addressed. This section evaluates framework robustness to localization inaccuracies by training separate instances under different user position error levels $\sigma_{\text{error}} \in \{0.0, 0.1, 0.3, 0.5, 1.0, 2.0, 3.0\}$ meters using error-matched training. The error follows the Gausian distribution with zero-mean and $\sigma_{\text{error}}$-variance. In this setup, each instance learns under the same error statistics present during evaluation, reflecting practical scenarios where localization system characteristics are known in advance, enabling error-aware policy learning.

Performance degrades systematically with increasing localization error, as shown in Fig.~\ref{Figure:localization_error_robustness}, The no-error baseline achieves $-64.43$ dBm mean RSSI. As error increases, performance decreases gradually:
\begin{itemize}
    \item $\sigma = 0.1$ m yields $-64.83$ dBm (0.40 dB loss);
    \item $\sigma = 0.3$ m yields $-70.58$ dBm (6.15 dB loss);
    \item $\sigma = 0.5$ m yields $-77.17$ dBm (12.74 dB loss);
    \item $\sigma = 1.0$ m yields $-78.16$ dBm (13.73 dB loss);
    \item $\sigma = 2.0$ m yields $-83.72$ dBm (19.29 dB loss);
    \item $\sigma = 3.0$ m yields $-89.56$ dBm (25.13 dB loss).
\end{itemize}

From a practical deployment perspective, approximately 0.3 m (30 cm) localization accuracy maintains less than 6 dB performance degradation, a requirement achievable through commodity WiFi/BLE infrastructure. Conversely, achieving tighter 0.1 m (10 cm) accuracy via Ultra-Wideband (UWB) systems enables near-optimal performance with only negligible loss. These trade-offs enable system designers to select localization technology based on performance requirements and cost constraints. The variance analysis presented in Fig.~\ref{Figure:localization_error_robustness} (shaded regions) reveals stable performance with sub-meter errors, showing less than 3 dBm standard deviation at 0.5 m error, but increasing standard deviation at larger errors (5.5 dBm at 3.0 m). This QoS consistency requirement suggests limiting localization error below 1.0 m for reliable operation in dynamic deployment scenarios.

While the framework exhibits sensitivity to errors exceeding 1.0 m, it maintains robust performance within the sub-meter regime typical of modern indoor tracking systems. This accuracy requirement aligns with the capabilities of emerging technologies, such as Ultra-Wideband (UWB), which routinely deliver the necessary decimeter-level precision. Consequently, the approach is well-suited for deployment in next-generation smart environments where high-fidelity localization infrastructure is available.

\section{Conclusion}
\label{sec:conclusion}

This work addresses the scalability and overhead challenges of multi-reflector coordination in mmWave systems by introducing an HMARL framework. By decomposing the optimization problem into high-level allocation and low-level execution layers, the proposed approach eliminates the dependency on explicit CSI, relying instead on geometric priors and user localization. Experimental evaluation confirms that the hierarchical architecture not only outperforms centralized optimization baselines by up to 7.94 dBm but also exhibits superior scalability. As user density increases, the system exploits multi-user diversity to maintain sustained total power efficiency with minimal per-user QoS degradation. Crucially, the framework demonstrates practical robustness: it adapts effectively across varying reflector aperture sizes and maintains reliable operation within the sub-meter localization error regime ($\le 0.5$ m) typical of modern tracking infrastructure.

These findings establish that mechanically reconfigurable reflectors, controlled via hierarchical learning, offer a cost-effective and wideband alternative to electronic metasurfaces for indoor coverage enhancement. Future research will extend this paradigm to high-mobility scenarios with time-varying channels, investigate joint AP-reflector beam-coordination, and validate the mechanical dynamics through physical prototyping.

\section*{Acknowledgments}
This material is based upon work supported by the U.S. Department of Energy, Office of Science, Office of Advanced Scientific Computing Research, Early Career Research Program under Award Number DE-SC-0023957.

\bibliographystyle{IEEEtran}
\bibliography{IEEEabrv,./main_ref}

\vfill

\end{document}